\documentclass{article}
\usepackage[utf8]{inputenc}
\usepackage{amsmath}
\usepackage{amssymb}
\usepackage{amsthm}
\usepackage{esint}
\usepackage{dsfont}
\usepackage{setspace}
\usepackage{geometry}
\usepackage{graphicx}
\usepackage{algorithm}
\usepackage{algpseudocode}
\usepackage{bbm}
\usepackage{bigints}
\usepackage{url}
\usepackage[inline]{enumitem}
\usepackage{graphicx}\usepackage{subcaption}
\usepackage{caption}
\usepackage[style=apa, backend=biber]{biblatex}
\usepackage{csquotes}

\title{ViASNet: A Video Ad Saliency Network for Predicting Dynamic Saliency and Viewer Engagement}

\begin{document}

\begin{titlepage}
    \centering
    \vspace*{1in} 
    {\Huge\bfseries ViASNet: A Video Ad Saliency Network for Predicting Dynamic Saliency and Viewer Engagement}\par
    \vspace{0.5in}
    {\Large Jianping Ye} \\ Department of Mathematics, \\ University of Maryland, College Park, MD 20742, USA\par
    {\tt\small jpye00@umd.edu\\}
    \vspace{0.2in}
    {\Large Michel Wedel} \\ Robert H. Smith School of Business, \\University of Maryland, College Park, MD 20742, USA\par
    {\tt\small mwedel@umd.edu\\}
  \vspace{0.2in}
     \vfill 
    {\large \today}\par

\thispagestyle{empty}

\begin{flushleft}

\vspace{3mm}

\textbf{Acknowledgement }
The authors are grateful to Tim Zuidgeest, CEO of Unravel Research, for providing the video and eye-tracking data.
\end{flushleft}
		
\end{titlepage}

\singlespacing
 
\title{ViASNet: A Video Ad Saliency Network for Predicting Dynamic Saliency and Viewer Engagement}

\maketitle

\thispagestyle{empty}

\maketitle

\begin{abstract}

The digital media landscape has seen a pervasive shift toward short-form video advertising on TV, social media and e-commerce platforms. The present study focuses on deep saliency prediction for short-form video advertising. Deep saliency models have been used to generate predictions of human eye fixation patterns with the purpose of enhancing user interaction with digital technology and optimizing its design. For video ads, dynamic saliency maps capture where and when viewers are looking, revealing why video ads are effective, and how their content should be optimized. We develop and test a new deep dynamic saliency prediction model called ViASNet (Video Ad Saliency Network), which has an architecture founded on the 3D U-Net, and accommodates the influence of audio and the semantic meaning of scenes. We assess the model's performance on 151 video ads, each seen by about 20 viewers wile their eye movements were tracked, and explore the critical factors influencing model performance through ablation experiments. We calculate the entropy of the predicted saliency maps frame-by-frame as a diagnostic tool to identify ads and scenes that fail to engage viewers, and illustrate its use on test data of 15 unseen ads. Our study reveals that ad design and testing can be sped up considerably through automated systems built on deep saliency models such as ViASNet.

\end{abstract}

\newpage

\section{Introduction}
\label{sec:Intro}

The digital media landscape has seen a pervasive shift toward short-form video advertising on social media and e-commerce platforms: video advertising has become the dominant form that advertising companies use to promote products and services on TV, on digital media platforms such as YouTube, TikTok, Facebook and Instagram, and on ad-supported tiers of streaming platforms such as Hulu and Netflix \parencite{meng2024impact}. Spending on short-form video ads reached over 110 billion in 2025 \parencite{nash2025adstats}. Across platforms, a wide variety of formats is deployed, ranging from bumper, shorts, pre-roll and mid-roll, in-stream, and in-feed video ads \parencite{Krishnan2013Understanding}.  While short-form video ads can achieve high user engagement -- the average time users are active on TikTok has more than doubled in the past five years \parencite{oberlo_tiktok_time} -- their omnipresence makes it  increasingly challenging to create ads that break through the clutter \parencite{meng2024impact}. 

\paragraph{Video ads' effectiveness.} Research has examined the effectiveness of video ads, frequently using neuro-physiological measurement tools such as eye tracking \parencite{ausin2021background, brasel2008breaking, brasel2008points, brasel2011media, brasel2014enhancing,janiszewski1993influence, teixeira2010moment, teixeira2012emotion, venkatraman2015predicting, wedel2008review}, facial recognition of emotions \parencite{liu2018video, teixeira2012emotion}, EEG \parencite{ausin2021background,barnett2017ticket,boksem2015brain,boksem2025eeg,christoforou2017your, venkatraman2015predicting}, and fMRI \parencite{guixeres2017consumer, hakim2021machines,tong2020brain, venkatraman2015predicting}. 

One stream of eye tracking literature in particular has focused  on the variability of eye fixations across viewers to assess ad effectiveness. \textcite{brasel2008points} found that viewers tend to fixate on the center of the screen when looking at TV ads and that the variability in frame-by-frame fixation locations was higher for TV commercials than for the TV shows in which the commercials were embedded, possibly reflecting a lower level of engagement. Related, \textcite{brasel2008breaking} found that during fast forwarding through commercials, viewers fixate on a small central area of the screen, which, if it contained the advetised brand, positively impacted attitudes, intentions, and behaviors. \textcite{teixeira2010moment} examined variability in ﬁxation locations for TV ads, and showed that a higher variability predicts ad avoidance (skipping or zapping). The authors suggested that higher variability in fixation locations indicates viewers' lack of  engagement with the video ad, and a failure of the ad's design to guide viewers' attention to key aspects of the scene in each frame. Further support for these conjectures comes from the research by \textcite{teixeira2012emotion}, who found relationships of the variability of fixation locations with viewers' emotions, measured from their facial expressions, and with ad avoidance. Thus, this prior literature has revealed that the variability of eye fixations provides measures of viewer engagement and ad effectiveness.  

\paragraph{Saliency maps.} Various measures of fixation variability have in common that they are computed from probabilistic maps of the fixation locations for each frame in a video, also called saliency maps. Saliency models are used to generate predictions of human eye movements, with the purpose of facilitating user interaction with consumer-facing technologies and optimizing their designs. In the marketing literature the estimation of static, probabilistic saliency maps was pioneered by \textcite{lans2008eye, Lans2008competitive, vanderlans2021online}. For video ads, dynamic saliency maps capture the time course of where viewers look, providing diagnostics on \textit{why} ads are effective, and \textit{which} content should be optimized. While early saliency methods like the Itti-Koch model \parencite{itti1998model} exploited only low-level visual features in static images, high-level semantic content exerts a major influence as well \parencite{dong2024short, xiao2023exploring}. In most recent work (see e.g. \nptextcite{Lans2021online}) saliency maps are derived from human eye movements, which are guided by a combination of low-level visual features (such as contrast, color, and motion; \nptextcite{lans2008eye}) and high-level semantic information (such as objects, faces, goals, and contexts; \nptextcite{Lans2008competitive}). 

\paragraph{Deep saliency models.} Deep saliency prediction models (see \nptextcite{borji2019saliency} for a review) hold the promise of automating the testing of video clips, dramatically reducing time and cost involved in data collection and processing. They have been applied not only to cinematic editing \parencite{moorthy2020gazed}, and video advertising \parencite{wang2024deep}, but also in technologies such as robotic guidance \parencite{butko2008visual}, human-robot interaction \parencite{schillaci2013evaluating}, video surveillance \parencite{yubing2011spatiotemporal}, autonomous driving \parencite{lateef2021saliency}, and human-computer interaction \parencite{jiang2023ueyes}.  Deep saliency models such as DeepGaze \parencite{kummerer2016deepgaze} and SalGAN \parencite{pan2017salgan}, trained on both image and eye-movement data, outperform classical models such as  \textcite{itti1998model} in saliency prediction on static images. Deep saliency models are currently the state-of-the-art for predicting when and where humans look in dynamic scenes \parencite{borji2019saliency}, but their performances still lag behind those for static scenes \parencite{djilali20203dsal}. 

Although saliency prediction has made enormous progress, dynamic saliency predictions remains a challenge, because of the computational difficulties of modeling the temporal dynamics of video frames and eye movements simultaneously. First, motions of objects attract human attention and therefore selectively impact the saliency of objects in each video frame, causing sparsity in fixation patterns and the corresponding dynamic saliency maps. Second, video ads comprise rapid scene changes that are purposefully designed to attract attention, which makes saliency tracking more difficult: dynamic saliency models need to account for these abrupt scene changes induced by scene cuts. Third, top-down information, specifically scene semantics, influences saliency, which may cause long-range dependencies in saliency across scenes. Fourth, video ads involve audio streams that may affect saliency maps by enhancing specific visual features. These features make dynamic saliency prediction for video ads inherently complex, necessitating novel model architectures. 

\paragraph{Goal of the present study.} In the present study, we \textit{develop a deep saliency prediction model for short-form video advertising, as a basis for assessing video ad effectiveness through measures of frame-by-frame fixation variability.} We develop an architecture, called ViASNet (Video Ad Saliency Network) tailored to deep saliency prediction of short-form video ads, accommodating the influence of scene cuts, semantic meaning of scenes, and audio. We explore the model's strengths and weaknesses; the critical factors influencing model performance are identified through ablation experiments that provide insights into how the architecture is optimized for deep saliency prediction. In section \ref{sec:Literature} we summarize prior research on neuro-physiological measurements of video ad effectiveness (section~\ref{subsec:neuro}), and static (section~\ref{subsec:stat_sal}) and dynamic (section~\ref{subsec:dynam_sal}) saliency prediction; in section~\ref{sec:data} we describe the video data used to calibrate and test our model, in section~\ref{sec:Method} we describe data preprocessing (section~\ref{subsec:preprocessing}) and the model architecture (sections~\ref{subsec:vision backbone} to \ref{subsec:centerbias}); in section~\ref{subsec:ablation_results} we describe ablation experiments, and in section~\ref{subsec:prediction_results} the predictive tests of the model. 

\section{Related Research}
\label{sec:Literature}

\subsection{Neurophysiological measurement of video ads}
\label{subsec:neuro}
Research has examined the effectiveness of video ads, often using neuro-physiological measurement tools such as eye tracking, EEG and fMRI. For example, \textcite{janiszewski1993influence} use eye tracking to show that attention to video ads can be enhanced via conditioning. \textcite{dydewalle1998film, germeys2007psychology} study the impact of cuts in movies on eye movements and find that viewers redirect attention to the most informative part on the next scene after a cut. \textcite{dydewalle1991watching, brasel2014enhancing} study how subtitles attract attention. \textcite{liu2018video} demonstrated that movie trailers can be reduced to short emotive clips using facial recognition of emotions. Both \textcite{brasel2011media} and \textcite{brasel2017media} studied multitasking and showed that breaks between TV commercials and TV shows leads viewers to switch towards viewing computer screens.  The research by \textcite{boksem2015brain}, \textcite{christoforou2017your}, \textcite{barnett2017ticket}, and \textcite{boksem2025eeg} reveals that EEG measures are useful to predict preferences, watching decisions, and commercial success of movie trailers; \textcite{guixeres2017consumer} and \textcite{hakim2021machines} show that they can be used to predict performance metrics of videos on YouTube, the latter research by using machine learning tools. \textcite{ausin2021background} use both eye tracking and EEG to study the effects of congruence of audio and video. \textcite{tong2020brain} use fMRI to predict attention to videos, while \textcite{venkatraman2015predicting} compare eye tracking, EEG, and fMRI and find the latter to provide the highest predictive power. As discussed previously, \textcite{brasel2008points,brasel2008breaking,teixeira2010moment, teixeira2012emotion} study the center bias of eye fixations and frame-to-frame fixation variability. All in all, this body of research has shown that the effectiveness of video ads can be rigorously tested via neuro-physiological measurement, and via eye tracking in particular. 

\subsection{Saliency perception}

Saliency is the degree to which regions in a scene are perceived to stand-out relative to their surroundings, based on visual features such as colors, orientation, contrast, size, and motion \parencite{wolfe2004attributes, treisman1980feature, borji2012state, peschel2013review,}. Saliency is represented in several areas in the visual brain as retinotopic spatial maps  \parencite{gottlieb1998representation, thompson2005visual}. The same areas/maps are also involved in guiding eye movements. Through winner-takes-all (WTA) and inhibition-of-return (IOR) selection processes \parencite{posner1984components, klein2000inhibition, theeuwes2005attentional}, the eyes are shifted from one salient region to the next \parencite{itti1998model, simola2011impact, vanderlans2021online}. In addition to these bottom-up processes, the eyes focus on features that are relevant to the viewer and the task and ignore those that are irrelevant, via top-down modulation of the saliency map \parencite{duncan1997competitive, zehetleitner2013salience,Lans2008competitive}, which involves the enhancement of task-relevant and inhibition of task-irrelevant features  \parencite{sawaki2010capture,vanderlans2021online}. 

\paragraph{Auditory input.} Auditory and visual input are integrated in some of the same brain areas that represent the visual saliency map \parencite{stein2004multisensory, wright2008orienting}. Auditory-visual integration has been shown to affect visual selection of specific features such as color, orientation, and motion \parencite{girelli1997same,staufenbiel2011spatially}, via their enhancement and inhibition \parencite{stein1996enhancement,noesselt2008sound}.

\paragraph{The time course of visual saliency.} Studies by \textcite{einhauser2008objects,donk2008effects,theeuwes2018visual,vanheusden2021dynamics} have shown that the effects of bottom-up saliency are short-lived, in part because top-down processes override them rather quickly.  \textcite{orquin2015effects} found the effects of saliency to be much more persistent. The studies in question, however, used static rather than dynamic stimuli. 

For dynamic scenes, it has been confirmed that visually salient regions are more likely to be fixated on \parencite{lemeur2007predicting,franchak2022age}. Editorial cuts have been shown to disrupt eye movements, resulting in different eye fixation patterns as compared to those observed during the viewing of natural scenes \parencite{dorr2010variability, dydewalle1998film,hirose2010perception,tatler2011eye}. In particular, viewers tend to reorient and fixate the center of the frame after a cut to a new shot \parencite{dydewalle1998film,lemeur2007predicting,mital2011clustering}. Research has also shown that the narrative coherence of video content induces consistency in eye movements across viewers \parencite{dorr2010variability,kirkorian2018effect, jing2023effect}, which is caused by viewers' anticipation of upcoming content \parencite{kirkorian2017anticipatory, jing2023effect}. \\

\subsection{Static saliency prediction}
\label{subsec:stat_sal}
Given the potential of saliency maps as a diagnostic tool to enhance user interaction with technology and
to optimize its design, a growing stream of research has developed prediction models for saliency maps. We briefly review the approaches here, extensive reviews are provided by \textcite{borji2012state, borji2019saliency}. Several classes of approaches have been used to predict saliency on static images: \textbf{computational, statistical, and deep learning}. 

\paragraph{Computational saliency models.} \textcite{itti1998model, itti2001computational} computed saliency for static images from low-level feature contrasts. Their saliency model was extended by including multi-resolution features \parencite{garcia2012saliency, goferman2011context, seo2009static}, peripheral vision \parencite{wang2017scanpath,wang2011simulating,wu2017saliency, wloka2018active}, center-surround mechanisms \parencite{kienzle2009center,tavakoli2013stochastic}, visual symmetry \parencite{kootstra2008paying}, gist \parencite{torralba2003modeling, torralba2006contextual, peters2007beyond}, regions of interest \parencite{gao2009discriminant, bruce2005saliency}, objects \parencite{walther2006modeling, rao2002eye}, surprise \parencite{itti2009bayesian}, prior knowledge \parencite{zhang2009sunday}, and top-down signals \parencite{deco2000hierarchical}. Boolean Map Saliency (BMS; \nptextcite{zhang2013saliency}) is based on figure-ground segmentation, and is one of the the best-performing computational models on MIT-T\"ubingen saliency benchmark. 

\paragraph{Statistical saliency models.} Statistical models cast the process by which saliency drives fixations in a probability framework. Research has used Markov Models \parencite{ellis1985patterns,pieters1999visual,salah2002selective,tavakoli2013stochastic},  Hidden Markov Models (HMM) \parencite{lans2008eye, pang2008stochastic, liu2013semantically}, and Hierarchical Bayes Models \parencite{Lans2008competitive}. Compared to computational models these probabilistic models offer the advantage of inference, testing, and prediction with a statistical likelihood-based framework.

\paragraph{Deep saliency models.} Generalizing the classical model by \textcite{itti1998model}, research has extended Deep Convolutional Neural Networks (CNN; \nptextcite{lecun1998gradient}) to allow for the prediction of fixation maps (see for a review \nptextcite{borji2019saliency}). Models build on a VGG-16 \parencite{simonyan2014very} backbone  \parencite{huang2015salicon, assens2018scanpath, kruthiventi2017deepfix, wang2017deep,cornia2016multi} or an AlexNet \parencite{krizhevsky2012imagenet} backbone \parencite{kummerer2014deep, kummerer2022deepgaze, huang2015salicon, pan2016shallow}.   Transformers \parencite{mondal2023gazeformer} and Large Language Models (LLM; \nptextcite{chen2025gazexplain}) include high-level semantic information to improve the prediction of various visual tasks \parencite{liu2016dhsnet, wang2018salient, zeng2019joint,lee2021railroad, chen2025gazexplain, byrne2023predicting, unger2024predicting}. \textcite{kummerer2014deep, kummerer2022deepgaze} show that deep saliency models outperform computational models \parencite{itti1998model, itti2001computational}. The DeepGaze family (DeepGaze I, II,  IIE, and III; \nptextcite{kummerer2014deep, kummerer2016deepgaze, linardos2021deepgaze,kummerer2022deepgaze}) has  saliency prediction capabilities with steadily improving accuracy. DeepGaze IIE ensembles multiple pretrained vision backbones (ResNet, DenseNet),  achieving state-of-the-art performance on the MIT-T\"ubingen saliency benchmark.  

\subsection{Dynamic saliency prediction}
\label{subsec:dynam_sal}

Initially, computational models for dynamic saliency prediction extended static saliency models with depth, motion and other dynamic features \parencite{itti2001computational, zhang2013saliency,leboran2016dynamic,leifman2017learning, rudoy2013learning}, but presently work on dynamic saliency prediction mostly involves deep learning models.  Several dynamic saliency models with CNN backbones have two separate modules that capture low-level spatial, respectively temporal/motion information \parencite{bak2017spatio, kocak2021gated, zhang2018video, sun2018sg}, or rely on static-to-dynamic transfer learning \parencite{chaabouni2016transfer}. Other models include Long-Short Term Memory models (LSTM; \nptextcite{hochreiter1997long}), other recurrent networks \parencite{bazzani2016recurrent, liu2017predicting, droste2020unified}, combined CNN-LSTM Networks \parencite{wang2018revisiting}, convolutional LSTM (ConvLSTM; \nptextcite{shi2015convolutional}) architectures \parencite{jiang2017predicting, gorji2018going}, or a diffusion model-based approach for audio-visual saliency \parencite{xiong2024diffsal}. Other recent models \parencite{zhou2023transformer, moradi2024transformer} capture long range dynamics using a transformer backbone \parencite{vaswani2017attention}. DeepGaze models such as DeepGaze IIE \parencite{linardos2021deepgaze} have been successfully applied to video saliency prediction on a frame-by-frame basis. 

\paragraph{3D CNNs.} 3D CNNs \parencite{ji20123d}, including UNet \parencite{ronneberger2015u}, have shown promise for dynamic saliency prediction \parencite{min2019tased,jain2021vinet,djilali20203dsal}, because they simultaneously capture spatial and temporal dimensions. 3DSal \parencite{djilali20203dsal} uses a VGG-16 backbone to extract features from consecutive frames and captures motion information via a 3D ConvNet architecture to predict the saliency map of the next frame. TASED-Net \parencite{min2019tased} uses a fully convolutional CNN, built on the pre-trained S3D \parencite{xie2018rethinking} motion recognition backbone. ViNet \parencite{jain2021vinet, girmaji2025minimalistic} extracts hierarchical features using a fully convolutional UNet-type encoder-decoder built on the S3D backbone \parencite{xie2018rethinking,pan2021actor}. \textcite{girmaji2025minimalistic} show that light-weight ViNet variants outperform most benchmark models. 
3DSal, TASED-Net and ViNet produce a single saliency map for the last frame only, but saliency maps for all frames can be predicted by applying the models repeatedly using a sliding-window.

\paragraph{Audio input.} Audio input has been incorporated by fusing auditory and visual feature embeddings via learned weights, attention mechanisms, or two-stream encoders \parencite{chang2021temporal,qiao2024joint,liu2020learning,xiong2023casp,xiong2024diffsal,jain2021vinet}. Several studies found, however, that incorporating audio information did not improve saliency predictions, relative to visual-only input \parencite{chang2021temporal,agrawal2022does,jain2021vinet,girmaji2025minimalistic}.

\paragraph{Conclusion.} Saliency prediction for static images has seen more extensive research than dynamic saliency prediction, where most recent work focuses on deep saliency models. State-of-the-art static saliency prediction models applied on a frame-to-frame basis to video cannot match the performance of dynamic saliency models, because they ignore temporal information. Several dynamic saliency prediction models, built on motion recognition backbones show state of the art performance on natural dynamic scenes, but are designed to predict the saliency map of the last video frame. For video ads, these dynamic saliency models for video ads do not allow for the inclusion of scene cuts and scene narratives, while it is currently not clear whether including audio enhances their effectiveness.

\section{Data}
\label{sec:data}
The dataset used for this research consists of 151 TV video ads that aired between  2023 and 2024 in the Netherlands. The data was collected by a professional research agency, Unravel Research\footnote{https://www.unravelresearch.com/en/home}. The video ads have an average length of about 30 seconds, and have 15.6 scenes on average, with an average scene length of 2 seconds. Each has an accompanying audio stream. Each video ad has 24 frames per second and on average there are 51.5 frames per scene. Each frame has a 1920 $\times$ 1080 pixel color resolution.

Each video ad was seen by 20 participants at the facilities of the research agency. Video ads were presented on a computer screen. The participants were instructed to watch the video ads as they would normally do at home. The raw gaze locations, $(x,y)$ pixel coordinates on each video frame, were collected with a pupil-center corneal reflection Tobii Pro Fusion\texttrademark eye tracker, with a sampling precision of 100 Hz, which provides about 1 sample of the gaze trajectory per 10 milliseconds. Fixations and saccades were identified from the raw gaze record with the Tobii algorithms,  which use an angular Identification-Velocity Threshold (IVT) of $30^\circ /\text{sec}$ to distinguish between low-velocity angular motion (fixations) and high-velocity angular motion (saccades) of the eye. Each frame of the video ad on average receives 179.5 eye fixations. These fixations include the $x-y$ coordinates of both left and right eyes. 
The data were split 80/20 for training and testing; the test data comprises 15 video ads. 


\section{Method}
\label{sec:Method}
\begin{figure}[!htb]
    \centering
    \includegraphics[width=1.0\linewidth]{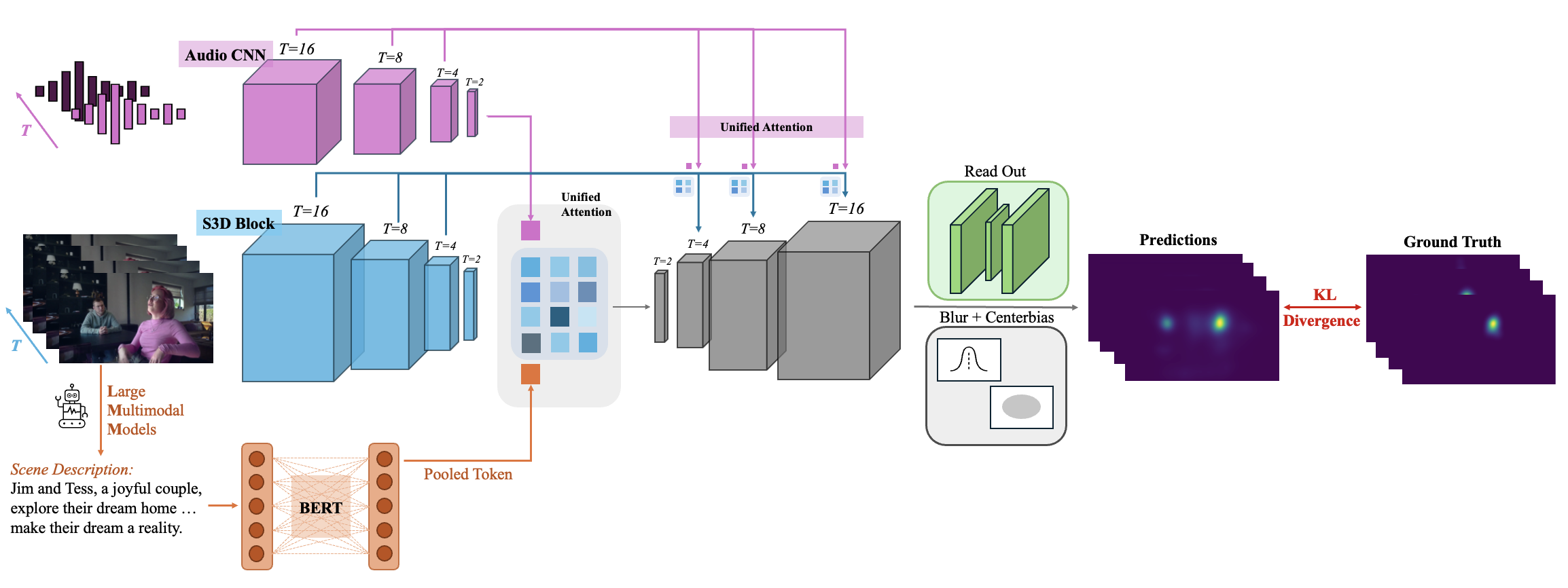}
    \caption{The diagram of ViASNet. ViASNet takes in three inputs: the visual frames of the target scene of the original video ad, the audio frames of the scene, and the scene descriptions generated by Qwen3-VL. ViASNet uses an S3D block, a customized audio CNN and the BERT-based text encoder to encode the multimodal features, and then integrates them by Unified Attention. At the final stage, the combined features are processed by an up-sampling decoder, followed by a trainable readout network, blurring, and a centerbias. ViASNet is trained end-to-end by the KL divergence between the predictions and the ground truth.}
    \label{fig:ViASNet Diagram}
\end{figure}

This section proposes ViASNet, \textbf{Vi}deo \textbf{A}d \textbf{S}aliency \textbf{Net}work, for end-to-end scene-wise ad video saliency predictions. ViASNet has a UNet-type structure \parencite{ronneberger2015u}, comprising a CNN encoder with a sequence of downsampling layers for audio and video, that are concatenated with embeddings of scene semantics via attention mechanisms, and a CNN decoder with a sequence of upsampling layers that have lateral connections to the downsampling layers, followed by a readout network to which blurring and centerbias mechanisms are added to predict the ground-truth saliency maps. Each of the following subsections explains one essential component of ViASNet. Figure \ref{fig:ViASNet Diagram} illustrates the overview of the network.

\subsection{Preprocessing}
\label{subsec:preprocessing}
Video ads involve “cuts” \parencite{liu2018video, dydewalle1998film}: discontinuous transitions between shots, or scenes. These scenes are the fundamental building blocks of video editing and help convey coherent narratives over time. We use PySceneDetect\footnote{\url{https://www.scenedetect.com}} to split each ad video into its scenes. We resize the frames to $224\times 384$ and normalize the pixels following \textcite{jain2021vinet}.

Rather than converting audio signals to the frequency domain using Fast Fourier Transforms, as is often done \parencite{abdel2014convolutional,zaman2023survey}, and to avoid subjective judgments and extensive pre-processing, we  use the raw audio signal as input to ViASNet. We align the audio frames with the visual frames. Then we normalize audio signals of each scene to the range $[-1,1]$. 

To obtain the ground-truth saliency map for each single frame for all videos, the steps in \textcite{bylinskii2018different} were followed. First, Gaussian convolution filters are applied to blur the fixations in each video frame. Frequency-based standard deviations are used for the gaussian filters \parencite{judd2009learning}. Then each frame is scaled down to size 224 $\times$ 384 (height $\times$ width) for computational efficiency. After rescaling, min-max normalization on each pixel converts the saliency maps to a common scale. These maps are transformed into a probability distribution over pixels on each frame by dividing each pixel value by the sum over all pixels.


\subsection{Vision Encoder}
\label{subsec:vision backbone}
ViASNet uses the pretrained S3D backbone \parencite{xie2018rethinking}, designed for motion recognition. The S3D architecture factorizes 3D convolutions into separate 2D spatial and 1D temporal convolutions. It uses a top-heavy design where lower layers utilize 2D convolutions, while upper layers utilize separable 3D convolutions. This prioritizes lower-level spatial feature extraction and upper-level temporal motion information, while reducing compute because the (factorized) 3D convolutions are applied to the smaller, abstract feature maps in the upper layers. S3D is applied to batches of scenes from each video ad; see section~\ref{subsec:preprocessing}. 

\subsection{Audio Encoder}
\label{subsec:audio processor}
The audio signals (section~\ref{subsec:preprocessing}) are processed by a customized CNN with 7 layers. It comprises a feature-pyramid architecture \parencite{lin2017feature} that produces multi-resolution feature maps by combining high-resolution features from lower layers with low-resolution semantic features from upper layers. The feature pyramid complements a vanilla bottom-up feed-forward 7-layer CNN with a top-down pathway that uses upsampling to reconstruct higher-resolution features from semantically rich upper layers. Lateral connections ($1 \times 1$ convolutions) merge features from the bottom-up pathway with the top-down pathway to recover the precise spatial locations lost in the feedforward pathway. 

\subsection{Scene Semantics Encoder}
\label{subsec:scene descriptor}
To describe the scene semantics in each cut, we apply the pre-trained Large Multimodal Model (LMM) Qwen3-VL-Plus \parencite{bai2025qwen3} to produce scene captions. To ensure consistency and capture the dynamics of semantic captions of adjacent scenes of the same video ad, at each target scene, we inject the caption of the previous scene in the prompt. Therefore, we prompt the LMM with: \textit{
This video is a scene from a TV commercial. Here is the caption of the previous scene: \{previous\_caption\}. Provide a caption describing the actions, objects, and scene in the current video.} 

Qwen3-VL-Plus cannot process clips shorter than 1 second. Therefore, for these scenes, we ask Qwen3-VL-Plus to first generate, respectively, three captions of the first, the middle and the last frames of the scene. Second, these captions are merged to produce the final caption for the scene. Captions of the previous scene are injected in the prompts fo each frame.

To produce semantics embeddings, we use a pretrained BERT (Bert-base-uncased \nptextcite{sanh2019distilbert}) interpreter of text descriptions of the scenes produced by the LMM. This BERT-based scene semantics encoder is fine-tuned during the training.

\subsection{Saliency Map Decoder}
\label{subsection: decoder}
ViASNet has a UNet-type structure \parencite{ronneberger2015u}, with a decoder that reconstructs the spatial resolution of the the audio-visual feature maps lost during downsampling, using a sequence of transposed convolutions. Each layer of distinct spatial dimensions in the decoder is connected to the corresponding layer in the encoder via lateral connections. These skip connections add encoder feature maps to corresponding decoder feature maps to preserve spatial details and improve training. The temporal upsampling is a simple trilinear interpolation, which follows standard spatial upsampling and does \textit{not} involve any parameters. 



\subsection{Multimodality Fusion}
\label{subsec:multimodality fusion}
The vision, audio, and semantic modules are connected by concatenating their feature maps, matching their dimensions through resizing and interpolation, and applying a Multimodal Unified Attention Network (MUAN) \parencite{yu2019multimodal} that  simultaneously captures intra- and inter-modal dependencies. Specifically, a Unified Attention Block processes visual, audio, and semantic features, allowing for intra-modal dependencies (audio-audio, video-video, semantic-semantic) via self-attention mechanisms, and for inter-modal dependencies (audio-video, audio-semantic, video-semantic) via co-attention mechanisms. We use a single MUAN block in the Unified Attention mechanism, which is applied after the concatenation of the multimodal embeddings, and again in each up-sampling step of the decoder. 

\subsection{Readout network}
\label{subsec:readout}
The readout network consists of a layer of $1 \times 1$ convolutions for cross-channel pooling that changes the feature map depth from 256 to 64, followed by $3 \times 3$ convolutions to capture spatial patterns, followed by a $1 \times 1$ transposed convolution to restore the original 256 channels. ReLU activations are applied to introduce nonlinearity. The readout network is similar to a ResNet bottleneck block \parencite{he2016deep}, and is much more parsimonious than a full $3 \times 3$ convolutional layer. 

\subsection{Centerbias and blur}
\label{subsec:centerbias}
The center bias component involves a spatial prior, modeled via a 2D Gaussian Kernel with trainable standard deviation, that is added to the network's output. Then the raw output network is blurred using a Gaussian convolution with a trainable standard deviation, as in DeepGaze \parencite{kummerer2022deepgaze}. A softmax activation is applied to the output of the spatial network to produce the saliency map, which represented as a spatial probability map over each frame in a video ad.

\section{Results}
\label{sec:results}
\subsection{Ablation Studies}
\label{subsec:ablation_results}
We investigate the contributions of the main components of ViASNet, that is \begin{enumerate*}[label=\arabic*)] \item the co-attention mechanism between the vision backbone and the Audio CNN, \item the co-attention mechanism between the vision backbone and the scene captions generated via the LMM, \item The unified co- and self-attention mechanisms across video, audio, and semantics, \item the readout network, and \item centerbias and blurring applied to the output of the readout network. \end{enumerate*} 

We use the following standard set of evaluation metrics \parencite{bylinskii2018different}: the Kullbach-Leibler distance (KL), the correlation coefficient (CC), the Normalized Scanpath Saliency (NSS), the $l_1$ distance (SIM), and the Area Under the Receiver Operator Curve, calculated according to Borji (AUC) and Judd (shuffled AUC; s-AUC) \parencite{kummererSaliencyBenchmarkingMade2018}.

\begin{table}[!htb]
\centering
\begin{tabular}{l|cccccc}
\hline
Model     & \multicolumn{1}{l}{\textit{KL} $\downarrow$} & \multicolumn{1}{l}{\textit{CC}}$\uparrow$ & \multicolumn{1}{l}{\textit{NSS}}$\uparrow$ & \multicolumn{1}{l}{\textit{SIM}} $\uparrow$ & \multicolumn{1}{l}{\textit{AUC}} $\uparrow$  & \multicolumn{1}{l}{\textit{s-AUC}} $\uparrow$ \\ \hline
Full              & \textbf{1.380}                      & \textbf{0.520}                  & \textbf{3.000}                   & \textbf{0.393}                   & \textbf{0.908}                   & \textbf{0.907}                     \\ \hline
$-$Audio Attention    & 1.415                               & 0.511                           & 2.962                            & 0.392                            & 0.907                            & 0.906                              \\
$-$Caption Attention & 1.400                               & 0.513                           & 2.928                            & 0.375                            & 0.907                            & 0.907                              \\
$-$Both Attention    & 1.435                               & 0.505                           & 2.890                            & 0.363                            & 0.905                            & 0.904                              \\
$-$Readout           & 1.433                               & 0.501                           & 2.873                            & 0.378                            & 0.906                            & 0.905                              \\
$-$Blurring          & 1.390                               & 0.517                           & 2.961                            & 0.390                            & 0.908                            & 0.907                              \\ \hline
\end{tabular}
\caption{Ablation studies of the components of ViASNet. The first row contains the results of the full model, and the subsequent rows respectively represent the results of each model without the component(s) specified. The best value in each column is indicated in bold. }
\label{tab:ablation_results}
\end{table}

Table~\ref{tab:ablation_results} shows first of all, that each component of the ViASNet adds to its performance, on all six measures (but improvements in AUC are small due to ceiling effects). Focusing on the KL divergence, the largest contribution comes from Unified Attention that merges the three modalities, which results in a 4.0\% increase in the KL when removed from the model, and the readout network, which results in a 3.8\% increase in the KL measure when zeroed out. Audio attention (2.5\%) and caption (semantics) attention (1.4\%) have smaller contributions, while blurring and center bias have the smallest contribution. Note that among all five model components, audio attention has the smallest effect on the NSS and SIM measures, but nonetheless has a measurable impact on saliency map predictions.   

\subsection{Prediction Results}
\label{subsec:prediction_results}
We next evaluate the performance of the full ViASNet on the test dataset; the results are shown in table~\ref{tab:prediction_results}. We benchmark ViASNet against four saliency models: 1. Itti-Koch \parencite{itti1998model, itti2001computational}; 2. BMS \parencite{zhang2013saliency}; 3. DeepGaze IIE \parencite{linardos2021deepgaze}; 4. ViNet-S \parencite{girmaji2025minimalistic}. Itti-Koch and BMS are computational saliency models to predict saliency maps bottom-up, the former being a standard baseline model and the latter the best performing model in its category. DeepGaze IIE and ViNet-S (an extension of ViNet) are deep learning models, the former trained on static images from MIT1003 (\cite{judd2009learning}) and the latter trained on movie clips from Hollywood-2 (\cite{6942210}). They are some of the best performing models deep saliency models for static, respectively, dynamic saliency prediction; we use their pre-trained versions. For Itti-Koch, BMS and DeepGaze IIE models, we apply them to each frame of the video clips and then measure their performance with the metrics listed in section \ref{subsec:ablation_results}. For ViNet-S, since it is trained on video clips of length 32 frames, we first augment our data by repeating the first frame at the front to match the desired length, as an inference preparation suggested by its authors.

ViASNet achieves excellent scores on all six dimensions, including best CC of 0.567, AUC of 0.914 and NSS of 3.016, which outperforms all baselines tested. Compared to the three frame-wise saliency prediction baselines, ViASNet significantly improves the performance by 20\% in AUC and over 50\% in other metrics over DeepGaze IIE, and even more over Itti \& Koch and BMS. Noticeably, the gap between ViASNet and DeepGaze IIE reveals the substantial benefit of dynamic over static saliency predictions, and suggests that temporal attention mechanisms play a significant role. ViASNet also performs better than ViNet-S, the strongest baseline trained on 487k frames of movie clips of high quality, on all metrics except SIM. Overall, the results indicate a strong positive relationship between the predicted and ground-truth saliency maps, and reveal that our model identifies key areas of visual saliency on the video ads.

\begin{table}[!htb]

\centering
\begin{tabular}{l|cccccc}
\hline
Model   & \textit{KL Div} $\downarrow$ & \textit{CC} $\uparrow$ & \textit{NSS} $\uparrow$ & \textit{SIM} $\uparrow$ & \textit{AUC} $\uparrow$ & \textit{s-AUC} $\uparrow$ \\ \hline
Itti-Koch & 2.690 & 0.074 & 0.136 & 0.146 & 0.591 & 0.590\\
BMS & 3.009 & 0.093 & -35.041 & 0.158 & 0.618 & 0.617\\
DeepGaze IIE & 2.276 & 0.231 & 1.056 & 0.190 & 0.766 & 0.765\\
ViNet-S & \underline{1.252} & \underline{0.560} & \underline{2.924} & \textbf{0.443} & \underline{0.912} & \underline{0.911}\\
\hline
\textbf{ViASNet} & \textbf{1.200}           & \textbf{0.567}       & \textbf{3.016}        & \underline{0.419}      & \textbf{0.914}        & \textbf{0.913}          \\ 
\hline
\end{tabular}
\caption{Prediction performance of ViASNet on the test data. The bold values are the best scores in each column, and the underlined values are the second best scores.}
\label{tab:prediction_results}
\end{table}

\begin{figure}[!h]
    \centering
    \includegraphics[width=0.5\linewidth]{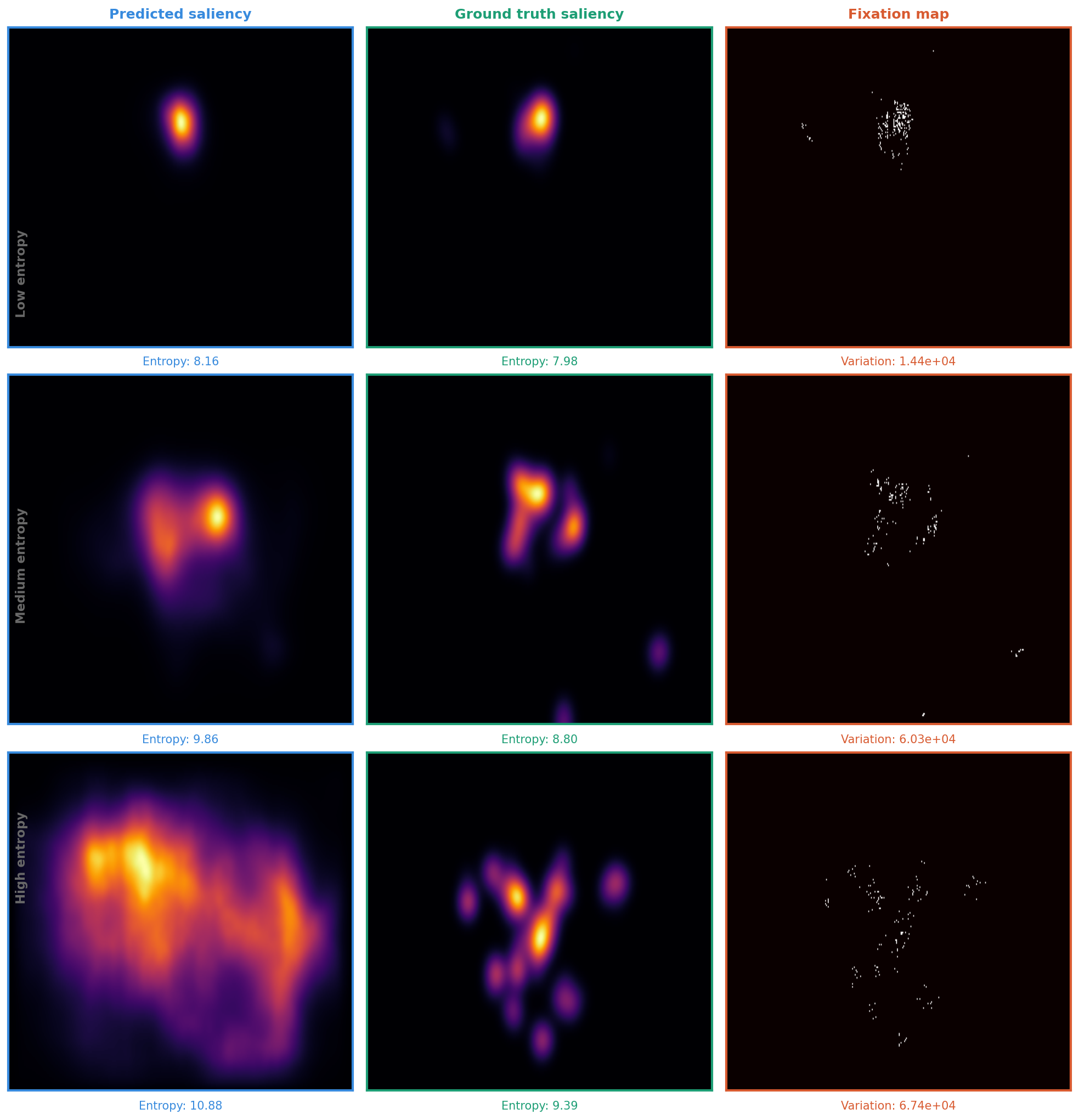}
    \caption{Saliency maps predicted by ViAsNet (left) for three non-cherry picked frames of a video in the test data, and corresponding ground-truth saliency maps (center) and fixation maps (right). The predicted entropy, ground truth entropy and fixation variability are shown in the subplot titles.}
    \label{fig:saliency_maps}
\end{figure}

Figure~\ref{fig:saliency_maps} shows the saliency maps predicted by ViAsNet (left) for three frames of low, medium and high predicted entropy in the test data from top to bottom rows, compared to the ground-truth saliency maps (center; see section \ref{subsec:preprocessing}), and the fixation maps (right). Their respective predicted entropy, ground truth entropy and observed fixation variability are all labeled in the titles of these subplots. The figure illustrates that ViASNet is does well in predicting saliency maps and reflecting the fixation variability, even when it does not have access to fixation data.

\subsection{Saliency Maps' Entropy}

As outlined in section \ref{sec:Intro}, prior research has shown that the frame-by-frame variability of eye fixations measures viewers' attentional focus and engagement. Because saliency maps predicted by ViASNet are probability maps (section \ref{subsec:centerbias}), we calculate, for the 15 video ads in the test data which were not seen by ViASNet, their frame-by-frame entropy as a measure of attentional focus. The entropy serves as a diagnostic tool for ad effectiveness. The hold-out correlation of the predicted vs. ground-truth entropy across all video ads/frames in the test data is 0.51. 

\begin{figure}[!h]
    \centering
    \includegraphics[width=0.8\linewidth]{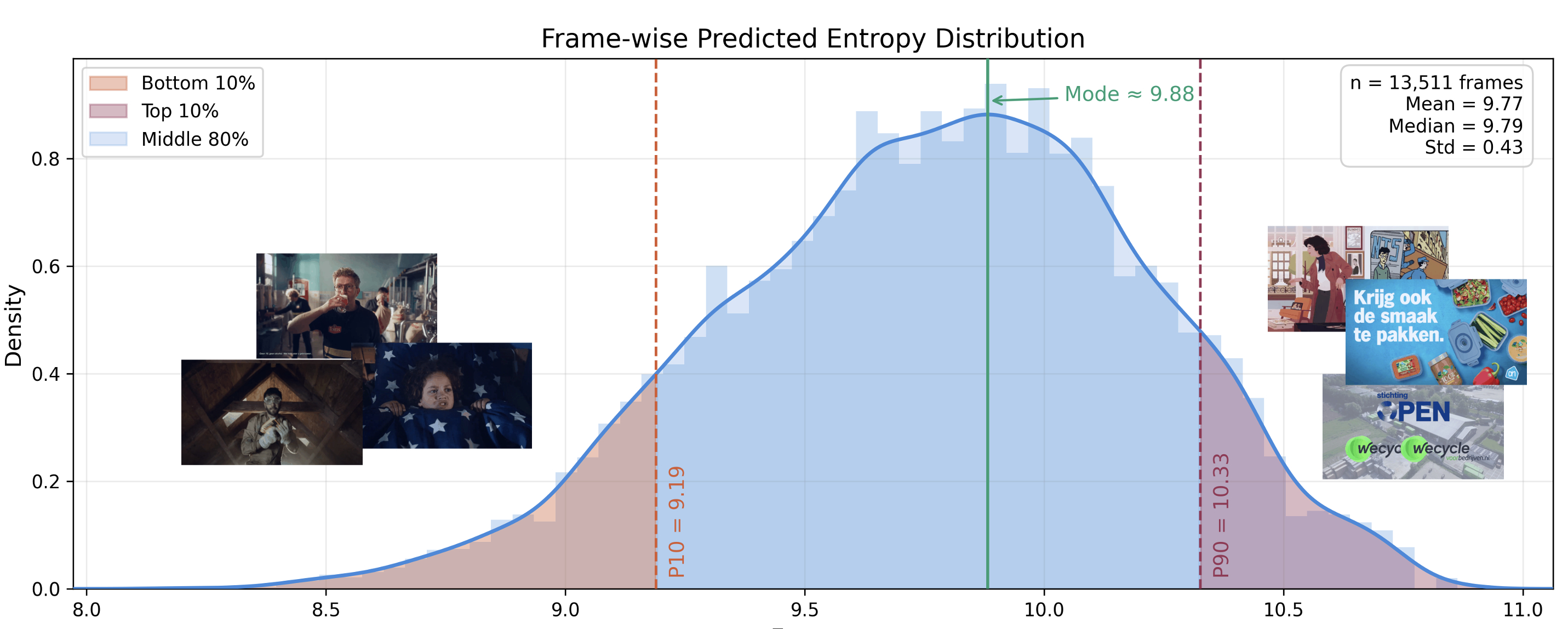}
    \caption{Empirical distribution of the entropy of the predicted saliency maps across all video frames in the test data, along with a few samples of frames in the $10^{th}$ (p10) and $90^{th}$ (p90) percentile of that distribution.}
    \label{fig:entropydistribution}
\end{figure}

Figure~\ref{fig:entropydistribution} shows the empirical distribution of the entropy of the predicted saliency maps across all frames for the test data, along with a few samples of frames in the tails of that distribution. The distribution is slightly left-skewed, with its mean below the median and mode. Frames below the $10^\text{th}$ percentile (p10) with low entropy ($ < 9.2$) have a high predicted engagement. These frames tend to have a person or face or the advertised product in their center. Frames above the $90^\text{th}$ percentile (p90) with high entropy ($> 10.3$) have low predicted engagement. These frames tend to have multiple objects and/or text spread across the frame (see figure~\ref{fig:entropydistribution}). Especially the latter frames above the p90 are of interest, because they could be improved or eliminated.  

\begin{figure}[!h]
    \centering
    \includegraphics[width=0.9\linewidth]{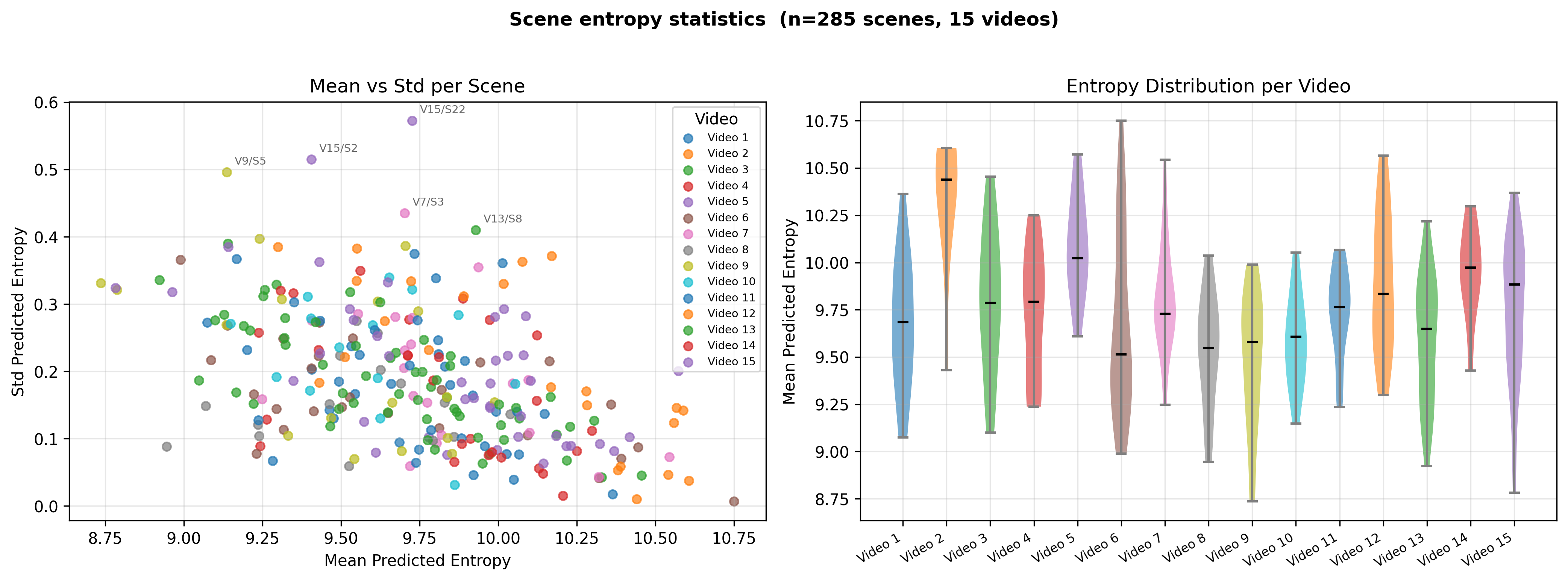}
    \caption{SD against the mean of the entropy for all scenes (left) and violin plots of the entropy for each video (right) in the test data.}
    \label{fig:entropy_scenes_ads}
\end{figure}

The left panel of Figure~\ref{fig:entropy_scenes_ads} plots the scene-specific SD of the entropy against its mean (across frames within the scene), for all scenes in all videos in the test data. The figure shows a strong negative relationship. This reveals that if a scene fails to capture attention (high mean), then that tends to be the case for all frames within the scene (low SD). Thus, when attention is lost, it is generally difficult to refocus viewers within that same scene: the entire scene doesn’t engage viewers. Further, this plot shows specifically which scenes/videos fail to engage viewers: for example, video 6 has one scene that falls above the p90 (figure \ref{fig:entropydistribution}) and is thus predicted to insufficiently capture attention of viewers; video 2 has a multiple scenes above the p90 that are predicted to insufficiently engage viewers. 

The right panel of Figure~\ref{fig:entropy_scenes_ads} shows violin plots of the entropy distribution for each of the 15 videos in the test data. These entropy distributions diagnose which videos are and which are not well designed to capture attention. For example videos 8, 9, 10 have low mean and relatively low entropy SD, and thus hold attention and are predicted to be engaging. Video 2, and to a lesser extent 5 and 14 aren’t as well designed and largely fall above the p90 (figure \ref{fig:entropydistribution}). Video 6 is generally engaging (low mean) but has one less engaging scene above the p90 (figure \ref{fig:entropydistribution}) that can be deleted, replaced, or improved (see also the plot in the left panel).

\begin{figure}[!h]
    \centering
    \includegraphics[width=0.9\linewidth]{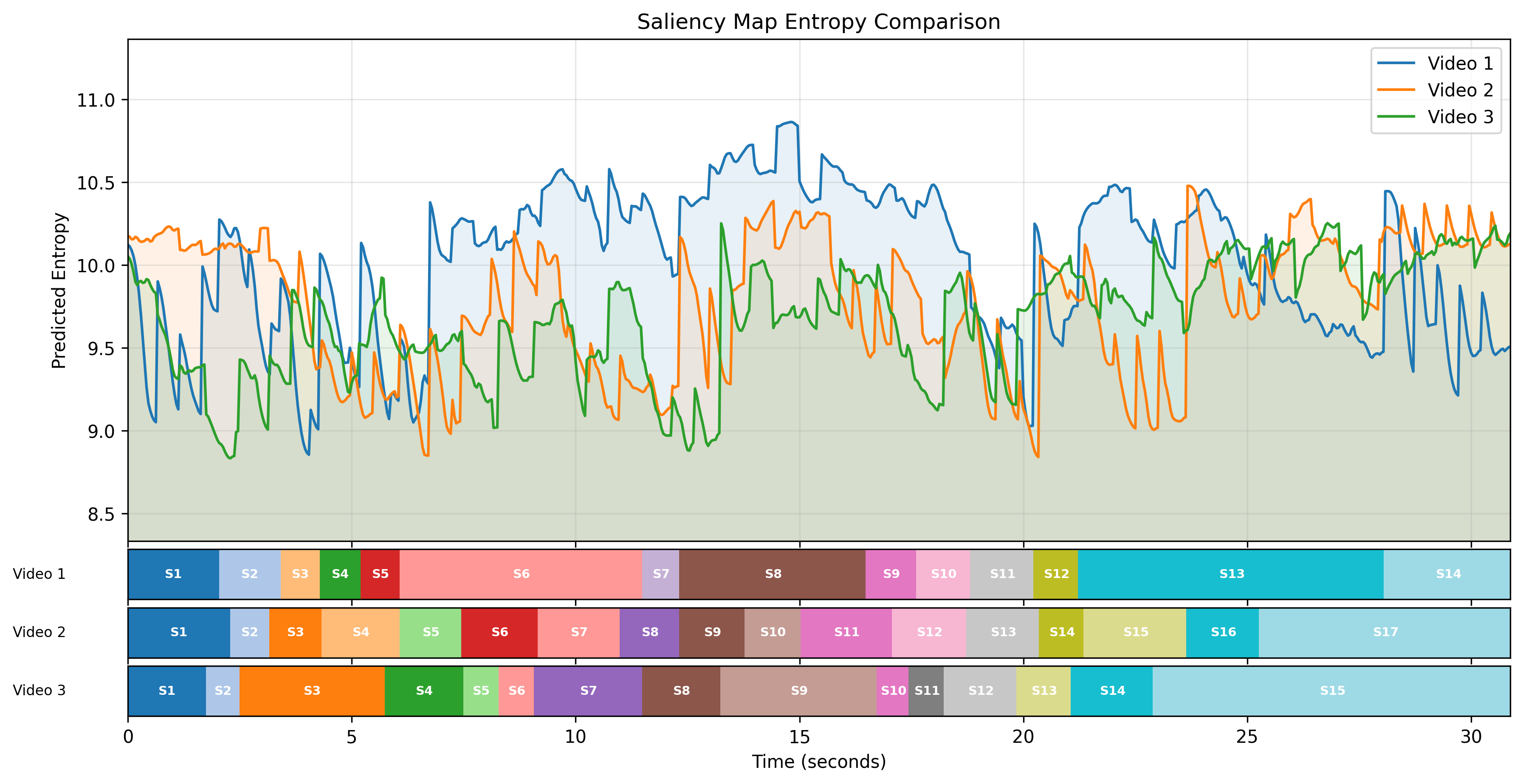}
    \caption{Frame-by-frame saliency map entropy progression plots for three sample videos (top), along with their scene composition (bottom).}
    \label{fig:entropy_dynamics}
\end{figure}

Finally, figure \ref{fig:entropy_dynamics} shows the frame-by-frame entropy progression for three video ads from the test data. Comparison of the frame-by-frame entropy with the p90 shows that while ad 3 is predicted to be engaging most of the time, video ad 1 is predicted to not focus viewers’ attention much of the time, while video 2 has a few scenes that fail to engage viewers. This plot provides detailed diagnostics for each video. For video ads 1 and 2 it helps identify the scenes that should be removed or improved: for video ad 1 those are the long scenes S6 and S8, and to some extent S13 (although engagement improves markedly over the course of that scene); video ad 2 is predicted to struggle to maintain attention at the very start (S1, S2), and the middle (S10, S16). At the end (S17), video ad 2 also tends to loose viewers' attentional focus, and video 3 has a similar problem.

Furthermore, it seems that in general scene cuts cause the entropy to jump up as viewers need to reorient themselves. That is, they cause a loss in viewers’ attention that needs to be regained after each cut.  The figure thus illustrates how ViASNet provides detailed diagnostics on videos' ability to capture viewers attention on a frame-by-frame basis, which, after it has been trained, is possible even without access to eye-movement data.

\section{Conclusion}
The shift toward short-form video advertising has resulted in high-volumes of low-cost video production on social media and other digital platforms, a trend that is accelerated by GenAI. Because short-form ads burn out more quickly, advertisers produce high volumes of them to engage consumers; AI may produce dozens of variations of a video ad simultaneously. In addition, individualized and customized video ad production is growing. These trends increase advertising clutter, making it difficult for ads to break through and capture attention, and  exacerbate challenges of rigorous video ad testing. The current state of affairs prioritizes rapid testing at scale, and hinders the application of eye tracking research, one of the industry standards in ad testing, because of the relatively high costs and efforts involved. Therefore, there is an increasing need for automated tools to analyze video ad performance which identifying ads and ad-scenes that do not engage viewers. ViASNet is a first step towards addressing that challenge. Calibrated on a large sample of video ads seen by consumers who's eye movements were tracked, ViASNet enables the prediction of dynamic saliency maps for unseen video ads, and the calculation of frame-by-frame saliency maps' entropy, as a measure of viewer engagement. ViASNet can be deployed as an automated testing tool to quickly identify ad variants that fail to capture viewers' attention. Especially for TV commercials with often higher production costs, it is a useful diagnostic tool to identify under-performing scenes that should be modified or removed. 

The strengths of ViASNet build on the recognition of the scene structure of video ads, the powerful UNet encoder-decoder architecture, and audio and semantics encoders that are integrated in the network via Unified Attention mechanisms. Future work could improve the architecture and its performance, build on real-time eye-tracking via webcams, integrate other data sources such as facial expressions of emotions, EEG, and incorporate the content of the advertising context. Rather than replacing human ad testing, we believe that ViASNet should be applied in Human-In-the-Loop (HITL) settings, combining the speed of automated testing with human capabilities to guide short-form video ad design and testing with strategic fit, authenticity, and ethics.

\newpage
\printbibliography 

@article{xiao2023exploring,
  title={Exploring the factors influencing consumer engagement behavior regarding short-form video advertising: A big data perspective},
  author={Xiao, Lin and Li, Xiaofeng and Zhang, Yucheng},
  journal={Journal of Retailing and Consumer Services},
  volume={70},
  pages={103170},
  year={2023},
  publisher={Elsevier}
}

@inproceedings{butko2008visual,
  title={Visual saliency model for robot cameras},
  author={Butko, Nicholas J and Zhang, Lingyun and Cottrell, Garrison W and Movellan, Javier R},
  booktitle={2008 IEEE International Conference on Robotics and Automation},
  pages={2398--2403},
  year={2008},
  organization={IEEE}
}

@article{lateef2021saliency,
  title={Saliency heat-map as visual attention for autonomous driving using generative adversarial network (GAN)},
  author={Lateef, Fahad and Kas, Mohamed and Ruichek, Yassine},
  journal={IEEE Transactions on Intelligent Transportation Systems},
  volume={23},
  number={6},
  pages={5360--5373},
  year={2021},
  publisher={IEEE}
}

@article{schillaci2013evaluating,
  title={Evaluating the effect of saliency detection and attention manipulation in human-robot interaction},
  author={Schillaci, Guido and Bodiro{\v{z}}a, Sa{\v{s}}a and Hafner, Verena Vanessa},
  journal={International Journal of Social Robotics},
  volume={5},
  number={1},
  pages={139--152},
  year={2013},
  publisher={Springer}
}

@inproceedings{moorthy2020gazed,
  title={Gazed--gaze-guided cinematic editing of wide-angle monocular video recordings},
  author={Moorthy, KL Bhanu and Kumar, Moneish and Subramanian, Ramanathan and Gandhi, Vineet},
  booktitle={Proceedings of the 2020 CHI Conference on Human Factors in Computing Systems},
  pages={1--11},
  year={2020}
}

@inproceedings{Krishnan2013Understanding,
author = {Krishnan, S. Shunmuga and Sitaraman, Ramesh K.},
title = {Understanding the effectiveness of video ads: a measurement study},
year = {2013},
isbn = {9781450319539},
publisher = {Association for Computing Machinery},
address = {New York, NY, USA},
url = {https://doi.org/10.1145/2504730.2504748},
doi = {10.1145/2504730.2504748},
booktitle = {Proceedings of the 2013 Conference on Internet Measurement Conference},
pages = {149–162},
numpages = {14},
location = {Barcelona, Spain},
series = {IMC '13}
}

@inproceedings{jiang2023ueyes,
  title={Ueyes: Understanding visual saliency across user interface types},
  author={Jiang, Yue and Leiva, Luis A and Rezazadegan Tavakoli, Hamed and RB Houssel, Paul and Kylm{\"a}l{\"a}, Julia and Oulasvirta, Antti},
  booktitle={Proceedings of the 2023 CHI conference on human factors in computing systems},
  pages={1--21},
  year={2023}
}

@article{wang2024deep,
  title={Deep learning-based saliency assessment model for product placement in video advertisements},
  author={Wang, Shaobo and Hu, Chenyu and Jia, Guancong and others},
  journal={Journal of Advanced Computing Systems},
  volume={4},
  number={5},
  pages={27--41},
  year={2024}
}

@article{borji2019saliency,
  title={Saliency prediction in the deep learning era: Successes and limitations},
  author={Borji, Ali},
  journal={IEEE transactions on pattern analysis and machine intelligence},
  volume={43},
  number={2},
  pages={679--700},
  year={2019},
  publisher={IEEE}
}

@article{liu2018video,
  title={Video content marketing: The making of clips},
  author={Liu, Xuan and Shi, Savannah Wei and Teixeira, Thales and Wedel, Michel},
  journal={Journal of Marketing},
  volume={82},
  number={4},
  pages={86--101},
  year={2018},
  publisher={SAGE Publications Sage CA: Los Angeles, CA}
}

@misc{nash2025adstats,
  Author={Nash, Bill},
  title = {Digital Advertising Statistics 2025},
  year = {2025},
  howpublished = {\url{https://marketingltb.com/blog/statistics/digital-advertising-statistics/}},
  note         = {Accessed: 2026-03-18}
}

@article{meng2024impact,
  title={The impact of content characteristics of Short-Form video ads on consumer purchase Behavior: Evidence from TikTok},
  author={Meng, Lu Monroe and Kou, Sining and Duan, Shen and Bie, Yongyue},
  journal={Journal of Business Research},
  volume={183},
  pages={114874},
  year={2024},
  publisher={Elsevier}
}

@article{dong2024short,
  title={Short video marketing: what, when and how short-branded videos facilitate consumer engagement},
  author={Dong, Xuebing and Liu, Hong and Xi, Nannan and Liao, Junyun and Yang, Zhi},
  journal={Internet Research},
  volume={34},
  number={3},
  pages={1104--1128},
  year={2024},
  publisher={Emerald Publishing Limited}
}

@misc{oberlo_tiktok_time,
  author       = {Oberlo},
  year = {2024},
  title        = {Average Time Spent on TikTok},
  howpublished = {\url{https://www.oberlo.com/statistics/average-time-spent-on-tiktok}},
  note         = {Accessed: 2024-11-15}
}

@article{itti1998model,
  title={A model of saliency-based visual attention for rapid scene analysis},
  author={Itti, Laurent and Koch, Christof and Niebur, Ernst},
  journal={IEEE Transactions on pattern analysis and machine intelligence},
  volume={20},
  number={11},
  pages={1254--1259},
  year={1998},
  publisher={Ieee}
}

@article{kummerer2016deepgaze,
  title={DeepGaze II: Reading fixations from deep features trained on object recognition},
  author={K{\"u}mmerer, Matthias and Wallis, Thomas SA and Bethge, Matthias},
  journal={arXiv preprint arXiv:1610.01563},
  year={2016}
}

@article{pan2017salgan,
  title={Salgan: Visual saliency prediction with generative adversarial networks},
  author={Pan, Junting and Ferrer, Cristian Canton and McGuinness, Kevin and O'Connor, Noel E and Torres, Jordi and Sayrol, Elisa and Giro-i-Nieto, Xavier},
  journal={arXiv preprint arXiv:1701.01081},
  year={2017}
}

@inproceedings{ronneberger2015u,
  title={U-net: Convolutional networks for biomedical image segmentation},
  author={Ronneberger, Olaf and Fischer, Philipp and Brox, Thomas},
  booktitle={Medical image computing and computer-assisted intervention--MICCAI 2015: 18th international conference, Munich, Germany, October 5-9, 2015, proceedings, part III 18},
  pages={234--241},
  year={2015},
  organization={Springer}
}

@inproceedings{jain2021vinet,
  title={Vinet: Pushing the limits of visual modality for audio-visual saliency prediction},
  author={Jain, Samyak and Yarlagadda, Pradeep and Jyoti, Shreyank and Karthik, Shyamgopal and Subramanian, Ramanathan and Gandhi, Vineet},
  booktitle={2021 IEEE/RSJ International Conference on Intelligent Robots and Systems (IROS)},
  pages={3520--3527},
  year={2021},
  organization={IEEE}
}

@inproceedings{wang2018revisiting,
  title={Revisiting video saliency: A large-scale benchmark and a new model},
  author={Wang, Wenguan and Shen, Jianbing and Guo, Fang and Cheng, Ming-Ming and Borji, Ali},
  booktitle={Proceedings of the IEEE Conference on computer vision and pattern recognition},
  pages={4894--4903},
  year={2018}
}

@article{ji20123d,
  title={3D convolutional neural networks for human action recognition},
  author={Ji, Shuiwang and Xu, Wei and Yang, Ming and Yu, Kai},
  journal={IEEE transactions on pattern analysis and machine intelligence},
  volume={35},
  number={1},
  pages={221--231},
  year={2012},
  publisher={IEEE}
}

@article{bylinskii2018different,
  title={What do different evaluation metrics tell us about saliency models?},
  author={Bylinskii, Zoya and Judd, Tilke and Oliva, Aude and Torralba, Antonio and Durand, Fr{\'e}do},
  journal={IEEE transactions on pattern analysis and machine intelligence},
  volume={41},
  number={3},
  pages={740--757},
  year={2018},
  publisher={IEEE}
}

@inproceedings{xie2018rethinking,
  title={Rethinking spatiotemporal feature learning: Speed-accuracy trade-offs in video classification},
  author={Xie, Saining and Sun, Chen and Huang, Jonathan and Tu, Zhuowen and Murphy, Kevin},
  booktitle={Proceedings of the European conference on computer vision (ECCV)},
  pages={305--321},
  year={2018}
}

@String(CVPR= {IEEE Conf. Comput. Vis. Pattern Recog.})

@String(ECCV= {Eur. Conf. Comput. Vis.})

@String(ICME = {Int. Conf. Multimedia and Expo})

@String(ICASSP=	{ICASSP})

@String(ICIP = {IEEE Int. Conf. Image Process.})

@String(AAAI = {AAAI})

@String(CVPR  = {CVPR})

@String(ECCV  = {ECCV})

@String(ICME  =	{ICME})

@String(ICIP  = {ICIP})

@inproceedings{kummererSaliencyBenchmarkingMade2018,
    title = {Saliency Benchmarking Made Easy: Separating Models, Maps and Metrics},
    series = {Lecture Notes in Computer Science},
    shorttitle = {Saliency Benchmarking Made Easy},
    pages = {798--814},
    booktitle = {Computer Vision – {ECCV} 2018},
    publisher = {Springer International Publishing},
    author = {K{\"u}mmerer, Matthias and Wallis, Thomas S. A. and Bethge, Matthias},
    editor = {Ferrari, Vittorio and Hebert, Martial and Sminchisescu, Cristian and Weiss, Yair},
    date = {2018},
}

@inproceedings{mondal2023gazeformer,
  title={Gazeformer: Scalable, effective and fast prediction of goal-directed human attention},
  author={Mondal, Sounak and Yang, Zhibo and Ahn, Seoyoung and Samaras, Dimitris and Zelinsky, Gregory and Hoai, Minh},
  booktitle={Proceedings of the IEEE/CVF conference on computer vision and pattern recognition},
  pages={1441--1450},
  year={2023}
}

@article{krizhevsky2012imagenet,
  title={Imagenet classification with deep convolutional neural networks},
  author={Krizhevsky, Alex and Sutskever, Ilya and Hinton, Geoffrey E},
  journal={Advances in Neural Information Processing Systems},
  volume={25},
  year={2012}
}

@article{vaswani2017attention,
  title={Attention is all you need},
  author={Vaswani, Ashish and Shazeer, Noam and Parmar, Niki and Uszkoreit, Jakob and Jones, Llion and Gomez, Aidan N and Kaiser, {\L}ukasz and Polosukhin, Illia},
  journal={Advances in Neural Information Processing Systems},
  volume={30},
  year={2017}
}

@article{lecun1998gradient,
  title={Gradient-based learning applied to document recognition},
  author={LeCun, Yann and Bottou, L{\'e}on and Bengio, Yoshua and Haffner, Patrick},
  journal={Proceedings of the IEEE},
  volume={86},
  number={11},
  pages={2278--2324},
  year={1998},
  publisher={Ieee}
}

@article{simonyan2014very,
  title={Very deep convolutional networks for large-scale image recognition},
  author={Simonyan, Karen and Zisserman, Andrew},
  journal={arXiv preprint arXiv:1409.1556},
  year={2014}
}

@article{kummerer2022deepgaze,
  title={DeepGaze III: Modeling Free-viewing Human Scanpaths with Deep Learning},
  author={K{\"u}mmerer, Matthias and Bethge, Matthias and Wallis, Thomas SA},
  journal={Journal of Vision},
  volume={22},
  number={5},
  pages={7--7},
  year={2022},
  publisher={The Association for Research in Vision and Ophthalmology}
}

@article{lans2008eye,
  title={Eye-Movement Analysis of Search Effectiveness},
  author={Van der Lans, Ralf and Pieters, Rik and Wedel, Michel},
  journal={Journal of the American Statistical Association},
  volume={103},
  number={482},
  pages={452--461},
  year={2008},
  publisher={Taylor \& Francis}
}

@article{hochreiter1997long,
  title={Long Short-Term Memory},
  author={Hochreiter, Sepp and Schmidhuber, J{\"u}rgen},
  journal={Neural Computation},
  volume={9},
  number={8},
  pages={1735--1780},
  year={1997},
  publisher={MIT press}
}

@inproceedings{he2016deep,
  title={Deep residual learning for image recognition},
  author={He, Kaiming and Zhang, Xiangyu and Ren, Shaoqing and Sun, Jian},
  booktitle={Proceedings of the IEEE conference on computer vision and pattern recognition},
  pages={770--778},
  year={2016}
}

@article{itti2001computational,
  title={Computational modelling of visual attention},
  author={Itti, Laurent and Koch, Christof},
  journal={Nature reviews neuroscience},
  volume={2},
  number={3},
  pages={194--203},
  year={2001},
  publisher={Nature Publishing Group UK London}
}

@article{kummerer2014deep,
  title={Deep gaze i: Boosting saliency prediction with feature maps trained on imagenet},
  author={K{\"u}mmerer, Matthias and Theis, Lucas and Bethge, Matthias},
  journal={arXiv preprint arXiv:1411.1045},
  year={2014}
}

@inproceedings{wang2018salient,
  title={Salient object detection driven by fixation prediction},
  author={Wang, Wenguan and Shen, Jianbing and Dong, Xingping and Borji, Ali},
  booktitle={Proceedings of the IEEE conference on computer vision and pattern recognition},
  pages={1711--1720},
  year={2018}
}

@inproceedings{zeng2019joint,
  title={Joint learning of saliency detection and weakly supervised semantic segmentation},
  author={Zeng, Yu and Zhuge, Yunzhi and Lu, Huchuan and Zhang, Lihe},
  booktitle={Proceedings of the IEEE/CVF international conference on computer vision},
  pages={7223--7233},
  year={2019}
}

@inproceedings{lee2021railroad,
  title={Railroad is not a train: Saliency as pseudo-pixel supervision for weakly supervised semantic segmentation},
  author={Lee, Seungho and Lee, Minhyun and Lee, Jongwuk and Shim, Hyunjung},
  booktitle={Proceedings of the IEEE/CVF conference on computer vision and pattern recognition},
  pages={5495--5505},
  year={2021}
}

@inproceedings{liu2016dhsnet,
  title={Dhsnet: Deep hierarchical saliency network for salient object detection},
  author={Liu, Nian and Han, Junwei},
  booktitle={Proceedings of the IEEE conference on computer vision and pattern recognition},
  pages={678--686},
  year={2016}
}

@inproceedings{chen2025gazexplain,
  title={GazeXplain: Learning to Predict Natural Language Explanations of Visual Scanpaths},
  author={Chen, Xianyu and Jiang, Ming and Zhao, Qi},
  booktitle={European Conference on Computer Vision},
  pages={314--333},
  year={2025},
  organization={Springer}
}

@article{unger2024predicting,
  title={Predicting consumer choice from raw eye-movement data using the RETINA deep learning architecture},
  author={Unger, Moshe and Wedel, Michel and Tuzhilin, Alexander},
  journal={Data Mining and Knowledge Discovery},
  volume={38},
  number={3},
  pages={1069--1100},
  year={2024},
  publisher={Springer}
}

@article{byrne2023predicting,
  title={Predicting choice behaviour in economic games using gaze data encoded as scanpath images},
  author={Byrne, Sean Anthony and Reynolds, Adam Peter Frederick and Biliotti, Carolina and Bargagli-Stoffi, Falco J and Polonio, Luca and Riccaboni, Massimo},
  journal={Scientific Reports},
  volume={13},
  number={1},
  pages={4722},
  year={2023},
  publisher={Nature Publishing Group UK London}
}

@article{klein2000inhibition,
  title={Inhibition of return},
  author={Klein, Raymond M},
  journal={Trends in cognitive sciences},
  volume={4},
  number={4},
  pages={138--147},
  year={2000},
  publisher={Elsevier}
}

@article{borji2012state,
  title={State-of-the-art in visual attention modeling},
  author={Borji, Ali and Itti, Laurent},
  journal={IEEE transactions on pattern analysis and machine intelligence},
  volume={35},
  number={1},
  pages={185--207},
  year={2012},
  publisher={IEEE}
}

@article{pieters1999visual,
  title={Visual attention to repeated print advertising: A test of scanpath theory},
  author={Pieters, Rik and Rosbergen, Edward and Wedel, Michel},
  journal={Journal of marketing research},
  volume={36},
  number={4},
  pages={424--438},
  year={1999},
  publisher={SAGE Publications Sage CA: Los Angeles, CA}
}

@inproceedings{judd2009learning,
  title={Learning to predict where humans look},
  author={Judd, Tilke and Ehinger, Krista and Durand, Fr{\'e}do and Torralba, Antonio},
  booktitle={2009 IEEE 12th international conference on computer vision},
  pages={2106--2113},
  year={2009},
  organization={IEEE}
}

@article{walther2006modeling,
  title={Modeling attention to salient proto-objects},
  author={Walther, Dirk and Koch, Christof},
  journal={Neural networks},
  volume={19},
  number={9},
  pages={1395--1407},
  year={2006},
  publisher={Elsevier}
}

@inproceedings{liu2013semantically,
  title={Semantically-based human scanpath estimation with HMMs},
  author={Liu, Huiying and Xu, Dong and Huang, Qingming and Li, Wen and Xu, Min and Lin, Stephen},
  booktitle={Proceedings of the IEEE international conference on computer vision},
  pages={3232--3239},
  year={2013}
}

@article{assens2018scanpath,
  title={Scanpath and saliency prediction on 360 degree images},
  author={Assens, Marc and Giro-i-Nieto, Xavier and McGuinness, Kevin and O’Connor, Noel E},
  journal={Signal Processing: Image Communication},
  volume={69},
  pages={8--14},
  year={2018},
  publisher={Elsevier}
}

@article{treisman1980feature,
  title={A feature-integration theory of attention},
  author={Treisman, Anne M and Gelade, Garry},
  journal={Cognitive psychology},
  volume={12},
  number={1},
  pages={97--136},
  year={1980},
  publisher={Elsevier}
}

@article{posner1984components,
  title={Components of visual orienting},
  author={Posner, Michael I and Cohen, Yoav and others},
  journal={Attention and performance X: Control of language processes},
  volume={32},
  pages={531--556},
  year={1984},
  publisher={Hilldale, NJ}
}

@article{wolfe2004attributes,
  title={What attributes guide the deployment of visual attention and how do they do it?},
  author={Wolfe, Jeremy M and Horowitz, Todd S},
  journal={Nature reviews neuroscience},
  volume={5},
  number={6},
  pages={495--501},
  year={2004},
  publisher={Nature Publishing Group UK London}
}

@article{thompson2005visual,
  title={A visual salience map in the primate frontal eye field},
  author={Thompson, Kirk G and Bichot, Narcisse P},
  journal={Progress in brain research},
  volume={147},
  pages={249--262},
  year={2005},
  publisher={Elsevier}
}

@article{Lans2021online,
  title={Online advertising suppresses visual competition during planned purchases},
  author={Van der Lans, Ralf and Pieters, Rik and Wedel, Michel},
  journal={Journal of Consumer Research},
  volume={48},
  number={3},
  pages={374--393},
  year={2021},
  publisher={Oxford University Press}
}

@article{deco2000hierarchical,
  title={A hierarchical neural system with attentional top--down enhancement of the spatial resolution for object recognition},
  author={Deco, Gustavo and Sch{\"u}rmann, Bernd},
  journal={Vision research},
  volume={40},
  number={20},
  pages={2845--2859},
  year={2000},
  publisher={Elsevier}
}

@article{wang2017scanpath,
  title={Scanpath estimation based on foveated image saliency},
  author={Wang, Yixiu and Wang, Bin and Wu, Xiaofeng and Zhang, Liming},
  journal={Cognitive processing},
  volume={18},
  number={1},
  pages={87--95},
  year={2017},
  publisher={Springer}
}

@inproceedings{wloka2018active,
  title={Active fixation control to predict saccade sequences},
  author={Wloka, Calden and Kotseruba, Iuliia and Tsotsos, John K},
  booktitle={Proceedings of the IEEE conference on computer vision and pattern recognition},
  pages={3184--3193},
  year={2018}
}

@article{torralba2006contextual,
  title={Contextual guidance of eye movements and attention in real-world scenes: the role of global features in object search.},
  author={Torralba, Antonio and Oliva, Aude and Castelhano, Monica S and Henderson, John M},
  journal={Psychological review},
  volume={113},
  number={4},
  pages={766},
  year={2006},
  publisher={American Psychological Association}
}

@inproceedings{wu2017saliency,
  title={Saliency map generation based on saccade target theory},
  author={Wu, Yue and Chen, Zhenzhong},
  booktitle={2017 IEEE International Conference on Multimedia and Expo (ICME)},
  pages={529--534},
  year={2017},
  organization={IEEE}
}

@article{ellis1985patterns,
  title={Patterns of statistical dependency in visual scanning},
  author={Ellis, Stephen R and Smith, James Darrell},
  journal={Eye movements and human information processing},
  pages={221--238},
  year={1985},
  publisher={Elsevier Amsterdam}
}

@inproceedings{wang2011simulating,
  title={Simulating human saccadic scanpaths on natural images},
  author={Wang, Wei and Chen, Cheng and Wang, Yizhou and Jiang, Tingting and Fang, Fang and Yao, Yuan},
  booktitle={CVPR 2011},
  pages={441--448},
  year={2011},
  organization={IEEE}
}

@article{tavakoli2013stochastic,
  title={Stochastic bottom--up fixation prediction and saccade generation},
  author={Tavakoli, Hamed Rezazadegan and Rahtu, Esa and Heikkil{\"a}, Janne},
  journal={Image and Vision Computing},
  volume={31},
  number={9},
  pages={686--693},
  year={2013},
  publisher={Elsevier}
}

@inproceedings{kootstra2008paying,
  title={Paying attention to symmetry},
  author={Kootstra, Gert and Nederveen, Arco and De Boer, Bart},
  booktitle={British Machine Vision Conference (BMVC2008)},
  pages={1115--1125},
  year={2008},
  organization={The British Machine Vision Association and Society for Pattern Recognition}
}

@article{torralba2003modeling,
  title={Modeling global scene factors in attention},
  author={Torralba, Antonio},
  journal={Journal of the Optical Society of America A},
  volume={20},
  number={7},
  pages={1407--1418},
  year={2003},
  publisher={OSA}
}

@article{itti2009bayesian,
  title={Bayesian surprise attracts human attention},
  author={Itti, Laurent and Baldi, Pierre},
  journal={Vision research},
  volume={49},
  number={10},
  pages={1295--1306},
  year={2009},
  publisher={Elsevier}
}

@inproceedings{zhang2009sunday,
  title={Sunday: Saliency using natural statistics for dynamic analysis of scenes},
  author={Zhang, Lingyun and Tong, Matthew H and Cottrell, Garrison W},
  booktitle={Proceedings of the 31st annual cognitive science conference},
  pages={2944--2949},
  year={2009},
  organization={AAAI Press Cambridge, MA}
}

@article{gao2009discriminant,
  title={Discriminant saliency, the detection of suspicious coincidences, and applications to visual recognition},
  author={Gao, Dashan and Han, Sunhyoung and Vasconcelos, Nuno},
  journal={IEEE Transactions on Pattern Analysis and Machine Intelligence},
  volume={31},
  number={6},
  pages={989--1005},
  year={2009},
  publisher={IEEE}
}

@article{bruce2005saliency,
  title={Saliency based on information maximization},
  author={Bruce, Neil and Tsotsos, John},
  journal={Advances in neural information processing systems},
  volume={18},
  year={2005}
}

@article{salah2002selective,
  title={A selective attention-based method for visual pattern recognition with application to handwritten digit recognition and face recognition},
  author={Salah, Albert Ali and Alpaydin, Ethem and Akarun, Lale},
  journal={IEEE Transactions on Pattern Analysis and Machine Intelligence},
  volume={24},
  number={3},
  pages={420--425},
  year={2002},
  publisher={IEEE}
}

@inproceedings{pang2008stochastic,
  title={A stochastic model of selective visual attention with a dynamic Bayesian network},
  author={Pang, Derek and Kimura, Akisato and Takeuchi, Tatsuto and Yamato, Junji and Kashino, Kunio},
  booktitle={2008 IEEE International Conference on Multimedia and Expo},
  pages={1073--1076},
  year={2008},
  organization={IEEE}
}

@article{rao2002eye,
  title={Eye movements in iconic visual search},
  author={Rao, Rajesh PN and Zelinsky, Gregory J and Hayhoe, Mary M and Ballard, Dana H},
  journal={Vision research},
  volume={42},
  number={11},
  pages={1447--1463},
  year={2002},
  publisher={Elsevier}
}

@article{kienzle2009center,
  title={Center-surround patterns emerge as optimal predictors for human saccade targets},
  author={Kienzle, Wolf and Franz, Matthias O and Sch{\"o}lkopf, Bernhard and Wichmann, Felix A},
  journal={Journal of vision},
  volume={9},
  number={5},
  pages={7--7},
  year={2009},
  publisher={The Association for Research in Vision and Ophthalmology}
}

@inproceedings{peters2007beyond,
  title={Beyond bottom-up: Incorporating task-dependent influences into a computational model of spatial attention},
  author={Peters, Robert J and Itti, Laurent},
  booktitle={2007 IEEE conference on computer vision and pattern recognition},
  pages={1--8},
  year={2007},
  organization={IEEE}
}

@article{Lans2008competitive,
  title={Research note—Competitive brand salience},
  author={Van der Lans, Ralf and Pieters, Rik and Wedel, Michel},
  journal={Marketing Science},
  volume={27},
  number={5},
  pages={922--931},
  year={2008},
  publisher={INFORMS}
}

@inproceedings{huang2015salicon,
  title={Salicon: Reducing the semantic gap in saliency prediction by adapting deep neural networks},
  author={Huang, Xun and Shen, Chengyao and Boix, Xavier and Zhao, Qi},
  booktitle={Proceedings of the IEEE international conference on computer vision},
  pages={262--270},
  year={2015}
}

@article{kruthiventi2017deepfix,
  title={Deepfix: A fully convolutional neural network for predicting human eye fixations},
  author={Kruthiventi, Srinivas SS and Ayush, Kumar and Babu, R Venkatesh},
  journal={IEEE Transactions on Image Processing},
  volume={26},
  number={9},
  pages={4446--4456},
  year={2017},
  publisher={IEEE}
}

@inproceedings{cornia2016multi,
  title={Multi-level net: A visual saliency prediction model},
  author={Cornia, Marcella and Baraldi, Lorenzo and Serra, Giuseppe and Cucchiara, Rita},
  booktitle={European Conference on Computer Vision},
  pages={302--315},
  year={2016},
  organization={Springer}
}

@inproceedings{pan2016shallow,
  title={Shallow and deep convolutional networks for saliency prediction},
  author={Pan, Junting and Sayrol, Elisa and Giro-i-Nieto, Xavier and McGuinness, Kevin and O'Connor, Noel E},
  booktitle={Proceedings of the IEEE conference on computer vision and pattern recognition},
  pages={598--606},
  year={2016}
}

@article{wang2017deep,
  title={Deep visual attention prediction},
  author={Wang, Wenguan and Shen, Jianbing},
  journal={IEEE Transactions on Image Processing},
  volume={27},
  number={5},
  pages={2368--2378},
  year={2017},
  publisher={IEEE}
}

@article{bak2017spatio,
  title={Spatio-temporal saliency networks for dynamic saliency prediction},
  author={Bak, Cagdas and Kocak, Aysun and Erdem, Erkut and Erdem, Aykut},
  journal={IEEE Transactions on Multimedia},
  volume={20},
  number={7},
  pages={1688--1698},
  year={2017},
  publisher={IEEE}
}

@inproceedings{chaabouni2016transfer,
  title={Transfer learning with deep networks for saliency prediction in natural video},
  author={Chaabouni, Souad and Benois-Pineau, Jenny and Amar, Chokri Ben},
  booktitle={2016 IEEE International Conference on Image Processing (ICIP)},
  pages={1604--1608},
  year={2016},
  organization={IEEE}
}

@article{jiang2017predicting,
  title={Predicting video saliency with object-to-motion CNN and two-layer convolutional LSTM},
  author={Jiang, Lai and Xu, Mai and Wang, Zulin},
  journal={arXiv preprint arXiv:1709.06316},
  year={2017}
}

@inproceedings{liu2017predicting,
  title={Predicting salient face in multiple-face videos},
  author={Liu, Yufan and Zhang, Songyang and Xu, Mai and He, Xuming},
  booktitle={Proceedings of the ieee conference on computer vision and pattern recognition},
  pages={4420--4428},
  year={2017}
}

@article{bazzani2016recurrent,
  title={Recurrent mixture density network for spatiotemporal visual attention},
  author={Bazzani, Loris and Larochelle, Hugo and Torresani, Lorenzo},
  journal={arXiv preprint arXiv:1603.08199},
  year={2016}
}

@inproceedings{gorji2018going,
  title={Going from image to video saliency: Augmenting image salience with dynamic attentional push},
  author={Gorji, Siavash and Clark, James J},
  booktitle={Proceedings of the IEEE Conference on Computer Vision and Pattern Recognition},
  pages={7501--7511},
  year={2018}
}

@article{sun2018sg,
  title={SG-FCN: A motion and memory-based deep learning model for video saliency detection},
  author={Sun, Meijun and Zhou, Ziqi and Hu, Qinghua and Wang, Zheng and Jiang, Jianmin},
  journal={IEEE transactions on cybernetics},
  volume={49},
  number={8},
  pages={2900--2911},
  year={2018},
  publisher={IEEE}
}

@article{kocak2021gated,
  title={A gated fusion network for dynamic saliency prediction},
  author={Kocak, Aysun and Erdem, Erkut and Erdem, Aykut},
  journal={IEEE Transactions on Cognitive and Developmental Systems},
  volume={14},
  number={3},
  pages={995--1008},
  year={2021},
  publisher={IEEE}
}

@article{zhang2018video,
  title={Video saliency prediction based on spatial-temporal two-stream network},
  author={Zhang, Kao and Chen, Zhenzhong},
  journal={IEEE Transactions on Circuits and Systems for Video Technology},
  volume={29},
  number={12},
  pages={3544--3557},
  year={2018},
  publisher={IEEE}
}

@inproceedings{droste2020unified,
  title={Unified image and video saliency modeling},
  author={Droste, Richard and Jiao, Jianbo and Noble, J Alison},
  booktitle={European Conference on Computer Vision},
  pages={419--435},
  year={2020},
  organization={Springer}
}

@inproceedings{min2019tased,
  title={Tased-net: Temporally-aggregating spatial encoder-decoder network for video saliency detection},
  author={Min, Kyle and Corso, Jason J},
  booktitle={Proceedings of the IEEE/CVF International Conference on Computer Vision},
  pages={2394--2403},
  year={2019}
}

@inproceedings{girmaji2025minimalistic,
  title={Minimalistic video saliency prediction via efficient decoder \& spatio temporal action cues},
  author={Girmaji, Rohit and Jain, Siddharth and Beri, Bhav and Bansal, Sarthak and Gandhi, Vineet},
  booktitle={ICASSP 2025-2025 IEEE International Conference on Acoustics, Speech and Signal Processing (ICASSP)},
  pages={1--5},
  year={2025},
  organization={IEEE}
}

@inproceedings{pan2021actor,
  title={Actor-context-actor relation network for spatio-temporal action localization},
  author={Pan, Junting and Chen, Siyu and Shou, Mike Zheng and Liu, Yu and Shao, Jing and Li, Hongsheng},
  booktitle={Proceedings of the IEEE/CVF Conference on Computer Vision and Pattern Recognition},
  pages={464--474},
  year={2021}
}

@article{zhou2023transformer,
  title={Transformer-based multi-scale feature integration network for video saliency prediction},
  author={Zhou, Xiaofei and Wu, Songhe and Shi, Ran and Zheng, Bolun and Wang, Shuai and Yin, Haibing and Zhang, Jiyong and Yan, Chenggang},
  journal={IEEE Transactions on Circuits and Systems for Video Technology},
  volume={33},
  number={12},
  pages={7696--7707},
  year={2023},
  publisher={IEEE}
}

@article{moradi2024transformer,
  title={Transformer-based video saliency prediction with high temporal dimension decoding},
  author={Moradi, Morteza and Palazzo, Simone and Spampinato, Concetto},
  journal={arXiv preprint arXiv:2401.07942},
  year={2024}
}

@article{shi2015convolutional,
  title={Convolutional LSTM network: A machine learning approach for precipitation nowcasting},
  author={Shi, Xingjian and Chen, Zhourong and Wang, Hao and Yeung, Dit-Yan and Wong, Wai-Kin and Woo, Wang-chun},
  journal={Advances in neural information processing systems},
  volume={28},
  year={2015}
}

@article{chang2021temporal,
  title={Temporal-spatial feature pyramid for video saliency detection},
  author={Chang, Qinyao and Zhu, Shiping},
  journal={arXiv preprint arXiv:2105.04213},
  year={2021}
}

@article{qiao2024joint,
  title={Joint learning of audio--visual saliency prediction and sound source localization on multi-face videos},
  author={Qiao, Minglang and Liu, Yufan and Xu, Mai and Deng, Xin and Li, Bing and Hu, Weiming and Borji, Ali},
  journal={International Journal of Computer Vision},
  volume={132},
  number={6},
  pages={2003--2025},
  year={2024},
  publisher={Springer}
}

@inproceedings{liu2020learning,
  title={Learning to predict salient faces: A novel visual-audio saliency model},
  author={Liu, Yufan and Qiao, Minglang and Xu, Mai and Li, Bing and Hu, Weiming and Borji, Ali},
  booktitle={European conference on computer vision},
  pages={413--429},
  year={2020},
  organization={Springer}
}

@inproceedings{xiong2023casp,
  title={CASP-Net: Rethinking video saliency prediction from an audio-visual consistency perceptual perspective},
  author={Xiong, Junwen and Wang, Ganglai and Zhang, Peng and Huang, Wei and Zha, Yufei and Zhai, Guangtao},
  booktitle={Proceedings of the IEEE/CVF conference on computer vision and pattern recognition},
  pages={6441--6450},
  year={2023}
}

@inproceedings{xiong2024diffsal,
  title={Diffsal: Joint audio and video learning for diffusion saliency prediction},
  author={Xiong, Junwen and Zhang, Peng and You, Tao and Li, Chuanyue and Huang, Wei and Zha, Yufei},
  booktitle={Proceedings of the IEEE/CVF Conference on Computer Vision and Pattern Recognition},
  pages={27273--27283},
  year={2024}
}

@inproceedings{agrawal2022does,
  title={Does audio help in deep audio-visual saliency prediction models?},
  author={Agrawal, Ritvik and Jyoti, Shreyank and Girmaji, Rohit and Sivaprasad, Sarath and Gandhi, Vineet},
  booktitle={Proceedings of the 2022 International Conference on Multimodal Interaction},
  pages={48--56},
  year={2022}
}

@article{peschel2013review,
  title={A review of the findings and theories on surface size effects on visual attention},
  author={Peschel, Anne O and Orquin, Jacob L},
  journal={Frontiers in psychology},
  volume={4},
  pages={902},
  year={2013},
  publisher={Frontiers Media SA}
}

@article{gottlieb1998representation,
  title={The representation of visual salience in monkey parietal cortex},
  author={Gottlieb, Jacqueline P and Kusunoki, Makoto and Goldberg, Michael E},
  journal={Nature},
  volume={391},
  number={6666},
  pages={481--484},
  year={1998},
  publisher={Nature Publishing Group UK London}
}

@article{duncan1997competitive,
  title={Competitive brain activity in visual attention},
  author={Duncan, John and Humphreys, Glyn and Ward, Robert},
  journal={Current opinion in neurobiology},
  volume={7},
  number={2},
  pages={255--261},
  year={1997},
  publisher={Elsevier}
}

@article{zehetleitner2013salience,
  title={Salience-based selection: Attentional capture by distractors less salient than the target},
  author={Zehetleitner, Michael and Koch, Anja Isabel and Goschy, Harriet and M{\"u}ller, Hermann Joseph},
  journal={PLoS One},
  volume={8},
  number={1},
  pages={e52595},
  year={2013},
  publisher={Public Library of Science San Francisco, USA}
}

@article{sawaki2010capture,
  title={Capture versus suppression of attention by salient singletons: Electrophysiological evidence for an automatic attend-to-me signal},
  author={Sawaki, Risa and Luck, Steven J},
  journal={Attention, Perception, \& Psychophysics},
  volume={72},
  number={6},
  pages={1455--1470},
  year={2010},
  publisher={Springer}
}

@article{vanderlans2021online,
  title={Online advertising suppresses visual competition during planned purchases},
  author={Van der Lans, Ralf and Pieters, Rik and Wedel, Michel},
  journal={Journal of Consumer Research},
  volume={48},
  number={3},
  pages={374--393},
  year={2021},
  publisher={Oxford University Press}
}

@article{simola2011impact,
  title={The impact of salient advertisements on reading and attention on web pages.},
  author={Simola, Jaana and Kuisma, Jarmo and {\"O}{\"o}rni, Anssi and Uusitalo, Liisa and Hy{\"o}n{\"a}, Jukka},
  journal={Journal of Experimental Psychology: Applied},
  volume={17},
  number={2},
  pages={174},
  year={2011},
  publisher={American Psychological Association}
}

@article{theeuwes2005attentional,
  title={Attentional capture and inhibition (of return): The effect on perceptual sensitivity},
  author={Theeuwes, Jan and Chen, Chi Yong Donny},
  journal={Perception \& psychophysics},
  volume={67},
  number={8},
  pages={1305--1312},
  year={2005},
  publisher={Springer}
}

@article{donk2008effects,
  title={Effects of salience are short-lived},
  author={Donk, Mieke and Van Zoest, Wieske},
  journal={Psychological Science},
  volume={19},
  number={7},
  pages={733--739},
  year={2008},
  publisher={SAGE Publications Sage CA: Los Angeles, CA}
}

@article{vanheusden2021dynamics,
  title={The dynamics of saliency-driven and goal-driven visual selection as a function of eccentricity},
  author={van Heusden, Elle and Donk, Mieke and Olivers, Christian NL},
  journal={Journal of Vision},
  volume={21},
  number={3},
  pages={2--2},
  year={2021},
  publisher={The Association for Research in Vision and Ophthalmology}
}

@article{einhauser2008objects,
  title={Objects predict fixations better than early saliency},
  author={Einh{\"a}user, Wolfgang and Spain, Merrielle and Perona, Pietro},
  journal={Journal of vision},
  volume={8},
  number={14},
  pages={18--18},
  year={2008},
  publisher={The Association for Research in Vision and Ophthalmology}
}

@article{orquin2015effects,
  title={Effects of salience are both short-and long-lived},
  author={Orquin, Jacob L and Lagerkvist, Carl Johan},
  journal={Acta psychologica},
  volume={160},
  pages={69--76},
  year={2015},
  publisher={Elsevier}
}

@article{theeuwes2018visual,
  title={Visual selection: Usually fast and automatic; seldom slow and volitional},
  author={Theeuwes, Jan},
  journal={Journal of cognition},
  volume={1},
  number={1},
  pages={29},
  year={2018}
}

@article{tatler2011eye,
  title={Eye guidance in natural vision: Reinterpreting salience},
  author={Tatler, Benjamin W and Hayhoe, Mary M and Land, Michael F and Ballard, Dana H},
  journal={Journal of vision},
  volume={11},
  number={5},
  pages={5--5},
  year={2011},
  publisher={The Association for Research in Vision and Ophthalmology}
}

@article{hirose2010perception,
  title={Perception and memory across viewpoint changes in moving images},
  author={Hirose, Yoriko and Kennedy, Alan and Tatler, Benjamin W},
  journal={Journal of vision},
  volume={10},
  number={4},
  pages={2--2},
  year={2010},
  publisher={The Association for Research in Vision and Ophthalmology}
}

@article{dorr2010variability,
  title={Variability of eye movements when viewing dynamic natural scenes},
  author={Dorr, Michael and Martinetz, Thomas and Gegenfurtner, Karl R and Barth, Erhardt},
  journal={Journal of vision},
  volume={10},
  number={10},
  pages={28--28},
  year={2010},
  publisher={The Association for Research in Vision and Ophthalmology}
}

@article{noesselt2008sound,
  title={Sound increases the saliency of visual events},
  author={Noesselt, Toemme and Bergmann, Daniel and Hake, Maria and Heinze, Hans-Jochen and Fendrich, Robert},
  journal={Brain research},
  volume={1220},
  pages={157--163},
  year={2008},
  publisher={Elsevier}
}

@incollection{stein2004multisensory,
  title={Multisensory integration in single neurons of the midbrain},
  author={Stein, Barry E and Jiang, Wan and Stanford, Terrence R},
  year={2004},
  editor={Calvert, G. and Spence, C. and Stein, B.E.},
  booktitle={The handbook of multisensory processes},
  publisher={MIT press},
  address={Cambridge}
}

@article{stein1996enhancement,
  title={Enhancement of perceived visual intensity by auditory stimuli: A psychophysical analysis},
  author={Stein, Barry E and London, Nancy and Wilkinson, Lee K and Price, Donald D},
  journal={Journal of cognitive neuroscience},
  volume={8},
  number={6},
  pages={497--506},
  year={1996},
  publisher={MIT Press}
}

@article{girelli1997same,
  title={Are the same attentional mechanisms used to detect visual search targets defined by color, orientation, and motion?},
  author={Girelli, Massimo and Luck, Steven J},
  journal={Journal of Cognitive Neuroscience},
  volume={9},
  number={2},
  pages={238--253},
  year={1997},
  publisher={MIT Press One Rogers Street, Cambridge, MA 02142-1209, USA journals-info~…}
}

@article{staufenbiel2011spatially,
  title={Spatially uninformative sounds increase sensitivity for visual motion change},
  author={Staufenbiel, Sabine M and Van der Lubbe, Rob HJ and Talsma, Durk},
  journal={Experimental brain research},
  volume={213},
  number={4},
  pages={457--464},
  year={2011},
  publisher={Springer}
}

@book{wright2008orienting,
  title={Orienting of attention},
  author={Wright, Richard D and Ward, Lawrence M},
  year={2008},
  publisher={Oxford University Press}
}

@article{franchak2022age,
  title={Age differences in orienting to faces in dynamic scenes depend on face centering, not visual saliency},
  author={Franchak, John M and Kadooka, Kellan},
  journal={Infancy},
  volume={27},
  number={6},
  pages={1032--1051},
  year={2022},
  publisher={Wiley Online Library}
}

@article{mital2011clustering,
  title={Clustering of gaze during dynamic scene viewing is predicted by motion},
  author={Mital, Parag K and Smith, Tim J and Hill, Robin L and Henderson, John M},
  journal={Cognitive computation},
  volume={3},
  number={1},
  pages={5--24},
  year={2011},
  publisher={Springer}
}

@article{lemeur2007predicting,
  title={Predicting visual fixations on video based on low-level visual features},
  author={Le Meur, Olivier and Le Callet, Patrick and Barba, Dominique},
  journal={Vision research},
  volume={47},
  number={19},
  pages={2483--2498},
  year={2007},
  publisher={Elsevier}
}

@article{kirkorian2018effect,
  title={Effect of sequential video shot comprehensibility on attentional synchrony: A comparison of children and adults},
  author={Kirkorian, Heather L and Anderson, Daniel R},
  journal={Proceedings of the National Academy of Sciences},
  volume={115},
  number={40},
  pages={9867--9874},
  year={2018},
  publisher={National Academy of Sciences}
}

@article{kirkorian2017anticipatory,
  title={Anticipatory eye movements while watching continuous action across shots in video sequences: A developmental study},
  author={Kirkorian, Heather L and Anderson, Daniel R},
  journal={Child development},
  volume={88},
  number={4},
  pages={1284--1301},
  year={2017},
  publisher={Oxford University Press}
}

@article{jing2023effect,
  title={The effect of narrative coherence and visual salience on children’s and adults’ gaze while watching video},
  author={Jing, Mengguo and Kadooka, Kellan and Franchak, John and Kirkorian, Heather L},
  journal={Journal of Experimental Child Psychology},
  volume={226},
  pages={105562},
  year={2023},
  publisher={Elsevier}
}

@article{djilali20203dsal,
  title={3dsal: An efficient 3d-cnn architecture for video saliency prediction},
  author={Djilali, Yasser Abdelaziz Dahou and Sayah, Mohamed and McGuinness, Kevin and O'Connor, Noel E},
  year={2020},
  publisher={ScitePress}
}

@article{garcia2012saliency,
  title={Saliency from hierarchical adaptation through decorrelation and variance normalization},
  author={Garcia-Diaz, Ant{\'o}n and Fdez-Vidal, Xos{\'e} R and Pardo, Xos{\'e} M and Dosil, Raquel},
  journal={Image and Vision Computing},
  volume={30},
  number={1},
  pages={51--64},
  year={2012},
  publisher={Elsevier}
}

@article{goferman2011context,
  title={Context-aware saliency detection},
  author={Goferman, Stas and Zelnik-Manor, Lihi and Tal, Ayellet},
  journal={IEEE transactions on pattern analysis and machine intelligence},
  volume={34},
  number={10},
  pages={1915--1926},
  year={2011},
  publisher={Ieee}
}

@article{seo2009static,
  title={Static and space-time visual saliency detection by self-resemblance},
  author={Seo, Hae Jong and Milanfar, Peyman},
  journal={Journal of vision},
  volume={9},
  number={12},
  pages={15--15},
  year={2009},
  publisher={The Association for Research in Vision and Ophthalmology}
}

@inproceedings{leifman2017learning,
  title={Learning gaze transitions from depth to improve video saliency estimation},
  author={Leifman, George and Rudoy, Dmitry and Swedish, Tristan and Bayro-Corrochano, Eduardo and Raskar, Ramesh},
  booktitle={Proceedings of the IEEE international conference on computer vision},
  pages={1698--1707},
  year={2017}
}

@inproceedings{zhang2013saliency,
  title={Saliency detection: A boolean map approach},
  author={Zhang, Jianming and Sclaroff, Stan},
  booktitle={Proceedings of the IEEE international conference on computer vision},
  pages={153--160},
  year={2013}
}

@inproceedings{rudoy2013learning,
  title={Learning video saliency from human gaze using candidate selection},
  author={Rudoy, Dmitry and Goldman, Dan B and Shechtman, Eli and Zelnik-Manor, Lihi},
  booktitle={Proceedings of the IEEE Conference on Computer Vision and Pattern Recognition},
  pages={1147--1154},
  year={2013}
}

@article{leboran2016dynamic,
  title={Dynamic whitening saliency},
  author={Leboran, Victor and Garcia-Diaz, Anton and Fdez-Vidal, Xose R and Pardo, Xose M},
  journal={IEEE transactions on pattern analysis and machine intelligence},
  volume={39},
  number={5},
  pages={893--907},
  year={2016},
  publisher={IEEE}
}

@article{yubing2011spatiotemporal,
  title={A spatiotemporal saliency model for video surveillance},
  author={Yubing, Tong and Cheikh, Faouzi Alaya and Guraya, Fahad Fazal Elahi and Konik, Hubert and Tr{\'e}meau, Alain},
  journal={Cognitive Computation},
  volume={3},
  number={1},
  pages={241--263},
  year={2011},
  publisher={Springer}
}

@incollection{wedel2008review,
  title={A Review of Eye-Tracking Research in Marketing},
  author={Wedel, Michel and Pieters, Rik},
  booktitle = {Review of Marketing Research, Volume 4},
 publisher= {Emerald Group Publishing, Ltd.},
 pages={123--47},
 year={2008},
}

@article{brasel2008points,
  title={Points of view: Where do we look when we watch TV?},
  author={Brasel, S Adam and Gips, James},
  journal={Perception},
  volume={37},
  number={12},
  pages={1890--1894},
  year={2008},
  publisher={SAGE Publications Sage UK: London, England}
}

@article{brasel2008breaking,
  title={Breaking through fast-forwarding: Brand information and visual attention},
  author={Brasel, S Adam and Gips, James},
  journal={Journal of Marketing},
  volume={72},
  number={6},
  pages={31--48},
  year={2008},
  publisher={SAGE Publications Sage CA: Los Angeles, CA}
}

@article{teixeira2010moment,
  title={Moment-to-moment optimal branding in TV commercials: Preventing avoidance by pulsing},
  author={Teixeira, Thales S and Wedel, Michel and Pieters, Rik},
  journal={Marketing Science},
  volume={29},
  number={5},
  pages={783--804},
  year={2010},
  publisher={INFORMS}
}

@article{teixeira2012emotion,
  title={Emotion-induced engagement in internet video advertisements},
  author={Teixeira, Thales S and Wedel, Michel and Pieters, Rik},
  journal={Journal of marketing research},
  volume={49},
  number={2},
  pages={144--159},
  year={2012},
  publisher={SAGE Publications Sage CA: Los Angeles, CA}
}

@article{boksem2015brain,
  title={Brain responses to movie trailers predict individual preferences for movies and their population-wide commercial success},
  author={Boksem, Maarten AS and Smidts, Ale},
  journal={Journal of Marketing Research},
  volume={52},
  number={4},
  pages={482--492},
  year={2015},
  publisher={SAGE Publications Sage CA: Los Angeles, CA}
}

@article{brasel2011media,
  title={Media multitasking behavior: Concurrent television and computer usage},
  author={Brasel, S Adam and Gips, James},
  journal={Cyberpsychology, Behavior, and Social Networking},
  volume={14},
  number={9},
  pages={527--534},
  year={2011},
  publisher={SAGE Publications Sage CA: Los Angeles, CA}
}

@article{brasel2017media,
  title={Media multitasking: How visual cues affect switching behavior},
  author={Brasel, S Adam and Gips, James},
  journal={Computers in Human Behavior},
  volume={77},
  pages={258--265},
  year={2017},
  publisher={Elsevier}
}

@article{janiszewski1993influence,
  title={The influence of classical conditioning procedures on subsequent attention to the conditioned brand},
  author={Janiszewski, Chris and Warlop, Luk},
  journal={Journal of Consumer Research},
  pages={171--189},
  year={1993},
  publisher={JSTOR}
}

@article{brasel2014enhancing,
  title={Enhancing television advertising: same-language subtitles can improve brand recall, verbal memory, and behavioral intent},
  author={Brasel, S Adam and Gips, James},
  journal={Journal of the Academy of Marketing Science},
  volume={42},
  number={3},
  pages={322--336},
  year={2014},
  publisher={Springer}
}

@article{christoforou2017your,
  title={Your brain on the movies: a computational approach for predicting box-office performance from viewer’s brain responses to movie trailers},
  author={Christoforou, Christoforos and Papadopoulos, Timothy C and Constantinidou, Fofi and Theodorou, Maria},
  journal={Frontiers in neuroinformatics},
  volume={11},
  pages={72},
  year={2017},
  publisher={Frontiers Media SA}
}

@article{boksem2025eeg,
  title={Do EEG Metrics Derived from Trailers Predict the Commercial Success of Movies? A Systematic Analysis of Five Independent Datasets},
  author={Boksem, Maarten AS and van Diepen, Rosanne M and Eijlers, Esther and Boekel, Wouter and Smidts, Ale},
  journal={Journal of Marketing Research},
  volume={62},
  number={4},
  pages={703--720},
  year={2025},
  publisher={SAGE Publications Sage CA: Los Angeles, CA}
}

@article{tong2020brain,
  title={Brain activity forecasts video engagement in an internet attention market},
  author={Tong, Lester C and Acikalin, M Yavuz and Genevsky, Alexander and Shiv, Baba and Knutson, Brian},
  journal={Proceedings of the National Academy of Sciences},
  volume={117},
  number={12},
  pages={6936--6941},
  year={2020},
  publisher={National Academy of Sciences}
}

@article{barnett2017ticket,
  title={A ticket for your thoughts: Method for predicting content recall and sales using neural similarity of moviegoers},
  author={Barnett, Samuel B and Cerf, Moran},
  journal={Journal of Consumer Research},
  volume={44},
  number={1},
  pages={160--181},
  year={2017},
  publisher={Oxford University Press}
}

@article{guixeres2017consumer,
  title={Consumer neuroscience-based metrics predict recall, liking and viewing rates in online advertising},
  author={Guixeres, Jaime and Bign{\'e}, Enrique and Ausin Azofra, Jose M and Alcaniz Raya, Mariano and Colomer Granero, Adrian and Fuentes Hurtado, Felix and Naranjo Ornedo, Valery},
  journal={Frontiers in psychology},
  volume={8},
  pages={1808},
  year={2017},
  publisher={Frontiers Media SA}
}

@article{hakim2021machines,
  title={Machines learn neuromarketing: Improving preference prediction from self-reports using multiple EEG measures and machine learning},
  author={Hakim, Adam and Klorfeld, Shira and Sela, Tal and Friedman, Doron and Shabat-Simon, Maytal and Levy, Dino J},
  journal={International Journal of Research in marketing},
  volume={38},
  number={3},
  pages={770--791},
  year={2021},
  publisher={Elsevier}
}

@article{ausin2021background,
  title={The background music-content congruence of TV advertisements: A neurophysiological study},
  author={Ausin, Jose M and Bigne, Enrique and Mar{\'\i}n, Javier and Guixeres, Jaime and Alcaniz, Mariano},
  journal={European Research on Management and Business Economics},
  volume={27},
  number={2},
  pages={100154},
  year={2021},
  publisher={Elsevier}
}
\end{document}